\documentclass[letterpaper]{article} 
\usepackage[draft]{aaai25}  
\usepackage{times}  
\usepackage{helvet}  
\usepackage{courier}  
\usepackage[hyphens]{url}  
\usepackage{graphicx} 
\usepackage{natbib}  
\usepackage{caption} 
\usepackage{algorithm}
\usepackage{algorithmic}

\usepackage{newfloat}
\usepackage{listings}
\DeclareCaptionStyle{ruled}{labelfont=normalfont,labelsep=colon,strut=off} 
\floatstyle{ruled}
\newfloat{listing}{tb}{lst}{}
\floatname{listing}{Listing}
\newcommand{\modelname}{\texttt{AGFSync}}
\usepackage{amsmath} 
\usepackage{amssymb}
\usepackage{bm}
\usepackage{enumitem}
\usepackage{tcolorbox}
\usepackage{booktabs} 
\usepackage{makecell}
\usepackage[capitalise]{cleveref}
\newcommand{\PreserveBackslash}[1]{\let\temp=\\#1\let\\=\temp}
\newcolumntype{C}[1]{>{\PreserveBackslash\centering}p{#1}}
\newcolumntype{R}[1]{>{\PreserveBackslash\raggedleft}p{#1}}
\newcolumntype{L}[1]{>{\PreserveBackslash\raggedright}p{#1}}
\usepackage{threeparttable}
\usepackage{subcaption}
\usepackage{multirow}
\setlist[itemize,enumerate]{leftmargin=*, topsep=0pt}
\newsavebox\CBox
\def\textBF#1{\sbox\CBox{#1}\resizebox{\wd\CBox}{\ht\CBox}{\textbf{#1}}}

\title{\modelname{}: Leveraging AI-Generated Feedback for Preference Optimization in Text-to-Image Generation}
\author{
    Jingkun An\textsuperscript{\rm 1}\equalcontrib, Yinghao Zhu\textsuperscript{\rm 1}\equalcontrib, Zongjian Li\textsuperscript{\rm 2}\equalcontrib,  Enshen Zhou\textsuperscript{\rm 1}, \\
    Haoran Feng\textsuperscript{\rm 3}, Xijie Huang\textsuperscript{\rm 1}, Bohua Chen\textsuperscript{\rm 4}, Yemin Shi\textsuperscript{\rm 2}, Chengwei Pan\textsuperscript{\rm 1, \rm 5}\thanks{Corresponding author.}
}
\affiliations{
    \textsuperscript{\rm 1}Beihang University,
    \textsuperscript{\rm 2}Peking University,
    \textsuperscript{\rm 3}Tsinghua University,\\
    \textsuperscript{\rm 4}Huazhong University of Science and Technology,
    \textsuperscript{\rm 5}Zhongguancun Laboratory\\
    anjingkun02@gmail.com, pancw@buaa.edu.cn
}

\begin{document}

\maketitle

\begin{abstract}

Text-to-Image (T2I) diffusion models have achieved remarkable success in image generation. Despite their progress, challenges remain in both prompt-following ability, image quality and lack of high-quality datasets, which are essential for refining these models. As acquiring labeled data is costly, we introduce \modelname{}, a framework that enhances T2I diffusion models through Direct Preference Optimization (DPO) in a fully AI-driven approach. \modelname{} utilizes Vision-Language Models (VLM) to assess image quality across style, coherence, and aesthetics, generating feedback data within an AI-driven loop. By applying \modelname{} to leading T2I models such as SD v1.4, v1.5, and SDXL-base, our extensive experiments on the TIFA dataset demonstrate notable improvements in VQA scores, aesthetic evaluations, and performance on the HPSv2 benchmark, consistently outperforming the base models. \modelname{}'s method of refining T2I diffusion models paves the way for scalable alignment techniques. Our code and dataset are publicly available.

\begin{links}
  \link{Project}{https://anjingkun.github.io/AGFSync}
\end{links}

\end{abstract}

\section{Introduction}

The advent of Text-to-Image (T2I) generation technology represents a significant advancement in generative AI. Recent breakthroughs have predominantly utilized diffusion models to generate images from textual prompts~\cite{rombach2022highresolution,betker2023dalle3,podell2023sdxl,zhang2023adding, zhou2024minedreamer}.
However, achieving high fidelity and aesthetics in generated images poses challenges, including deviations from prompts and inadequate image quality~\cite{zhang2023text}. Addressing these challenges requires enhancing diffusion models' ability to accurately interpret detailed prompts (prompt-following ability~\cite{betker2023dalle3}) and improve the generative quality across style, coherence, and aesthetics.

Efforts to overcome these challenges span dataset, model, and training levels. High-quality text-image pair datasets, as proposed in the data-centric AI philosophy, can significantly improve performance~\cite{zhou2024lima}. Therefore a high-quality image caption and its corresponding image pair dataset is crucial in training~\cite{betker2023dalle3}.

At the model architecture level, advancements include the optimization of cross-attention mechanisms to improve model compliance~\cite{feng2023trainingfree}. These efforts, both at the dataset and model architecture levels, follow the traditional training paradigm of using elaborately designed models with specific datasets. In contrast, in the training domain, strategies inspired by the success of large language models, such as OpenAI's ChatGPT~\cite{blog2023ChatGPT}, include supervised finetuning (SFT) and alignment stages. With a pretrained T2I diffusion model, enhancing the model for better image quality can be approached in either the SFT stage or the alignment stage. The former approach, as seen in the latest work DreamSync~\cite{sun2023dreamsync}, finetunes the diffusion model through a selected image selection procedure where a Vision-Language Model (VLM)~\cite{openai2023gpt4,qin2023mp5,zhou2024code,qin2024worldsimbench} evaluates and then selects high-quality text-image pairings for further finetuning. However, DreamSync exhibits a lower prompt generation conversion rate and is limited by the intrinsic capabilities of the diffusion model, leading to uncontrollable data distribution in the finetuning dataset. The latter approach, DPOK~\cite{fan2024reinforcement}, DDPO~\cite{black2023trainingDDPO}, and DPO~\cite{rafailov2023DPO} use reinforcement learning for alignment, while Diffusion-DPO~\cite{wallace2023diffusionDPO} applies Direct Preference Optimization (DPO) for model alignment, modifying the original DPO algorithm to directly optimize diffusion models based on preference data. Yet, it only focuses on evaluating image quality from one aspect. Furthermore, existing methods mostly depend on extensive, quality-controlled labeled data.

Addressing these requires a cost-effective, low-labor approach that minimizes the need for human-labeled data while considering multiple quality aspects of images. Leveraging AI in generating datasets and evaluating image quality can fill these gaps without human intervention. Through generating diverse textual prompts, assessing generated images, and constructing a comprehensive preference dataset, \modelname{} epitomizes the full spectrum of AI-driven innovation—ushering in an era of enhanced data utility, accessibility, scalability, and process automation while simultaneously mitigating the costs and limitations associated with manual data labeling.

More specifically, \modelname{} aligns text-to-image diffusion models via DPO, with multi-aspect AI feedback generated data. The process begins with the preference candidate set generation, where LLM generates descriptions of diverse styles and categories, serving as high-quality textual prompts. Candidate images are then generated using these AI-generated prompts, therefore constructing candidate prompt-image pairs. Image evaluation and VQA data construction follow, using LLM to generate questions related to the composition elements, style, etc., based on its initial prompts. VQA scoring is conducted by inputting these questions into the VQA model to assess whether the diffusion model-generated images aesthetically follow the prompts, calculating accuracy as the VQA score. With combined weighted scores of VQA, CLIP, and aesthetics filtering, the preference pair dataset is established within the best and worst images. Finally, DPO alignment is applied to the diffusion model using the constructed preference pair dataset. The entire process leverages the robust capabilities of VLMs without any human engagement, ensuring a human-free, cost-effective workflow.

Our contributions are summarized as follows: 
\begin{enumerate}
 \item We introduce an openly accessible dataset composed of 45.8K AI-generated prompt samples and corresponding SDXL-generated images, each accompanied by question-answer pairs that validate the image generation's fidelity to textual prompts. This dataset not only propels forward the research in T2I generation but also embodies the shift towards higher data utilization, scalability, and generalization, signifying a breakthrough in mitigating the unsustainable practices of manual data annotation. 
  \item Our proposed framework \modelname{}, aided by multiple evaluation scores, leveraging DPO finetuning approach, introduces a fully automated, AI-driven approach, which elevates fidelity and aesthetic quality across varied scenarios without human annotations.  
  \item Extensive experiments demonstrate that \modelname{} significantly and consistently improves upon existing diffusion models in terms of adherence to text prompts and overall image quality, establishing the efficacy and transformative potential of our AI-driven data generation, evaluation and finetuning framework.
\end{enumerate}
\section{Related Work}

\subsection{Aligning Diffusion Models Methods}

The primary focus of related work in this area is to enhance the fidelity of images generated by diffusion models in response to text prompts, ensuring they align more closely with human preferences. This endeavor spans across dataset curation, model architecture enhancements, and specialized training methodologies.

\textbf{Dataset-Level Approaches}: A pivotal aspect of improving image generation models involves curating and finetuning datasets that are deemed visually appealing. Works by~\cite{podell2023sdxl,rombach2022highresolution} utilize datasets rated highly by aesthetics classifiers~\cite{schuhmann2022laionaes} to bias models towards generating visually appealing images. Similarly, Emu~\cite{dai2023emu} enhances both visual appeal and text alignment through finetuning on a curated dataset of high-quality photographs with detailed captions. Efforts to re-caption web-scraped image datasets for better text fidelity are evident in~\cite{betker2023dalle3,segalis2023picture}. Moreover, similar to finetuning LLMs with generated data~\cite{betker2023dalle3,segalis2023picture}, DreamSync~\cite{sun2023dreamsync} improves
T2I synthesis with feedback from vision-language image understanding models, aligning images with textual input and the aesthetic quality of the generated images.

\textbf{Model-Level Enhancements}: At the model level, enhancing the architecture with additional components like attention modules~\cite{feng2023trainingfree} offers a training-free solution to enhance model compliance with desired outputs. StructureDiffusion~\cite{feng2023trainingfree} and SynGen~\cite{rassin2024linguistic} also work on training-free methods that focus on model's inference time adjustments.

\textbf{Training-Level Strategies}: The integration of supervised finetuning (SFT) with advanced alignment stages, such as reinforcement learning approaches like DPOK~\cite{fan2024reinforcement}, DDPO~\cite{black2023trainingDDPO}, and DPO~\cite{rafailov2023DPO}, shows significant potential in aligning image quality with human preferences. Among these, Diffusion-DPO emerges as an RL-free method, distinct from other RL-based alignment strategies, effectively enhancing human appeal while ensuring distributional integrity~\cite{wallace2023diffusionDPO}.

A common drawback of these approaches is the expensive finetuning dataset, as most of them rely on human-annotated data and human evaluation. This paradigm cannot support training an extensive and scalable diffusion model.

\subsection{Image Quality Evaluation Methods}

Evaluating image quality in a comprehensive manner is pivotal, integrating both automated benchmarks and human assessments to ensure fidelity and aesthetic appeal. The introduction of TIFA~\cite{hu2023tifa} utilize Visual Question Answering (VQA) models to measure the faithfulness of generated images to text prompts, setting a foundation for subsequent innovations. The CLIP score~\cite{hessel2021clipscore} builds upon CLIP~\cite{radford2021clip} enables a reference-free evaluation of image-caption compatibility through the computation of cosine similarity between image and text embeddings, showcasing high correlation with human judgments without needing reference captions. PickScore~\cite{kirstain2024pick} leverages user preferences to predict the appeal of generated images, combining CLIP model elements with InstructGPT's reward model objectives~\cite{ouyang2022trainingInstructGPT} for a nuanced understanding of user satisfaction. Alongside, the aesthetic score~\cite{ke2023Aesthetics} assesses images based on aesthetics learned from image-comment pairs, providing a richer evaluation that includes composition, color, and style.

\section{Methodology}

\begin{figure*}[!ht]
  \centering
  \includegraphics[width=0.7\linewidth]{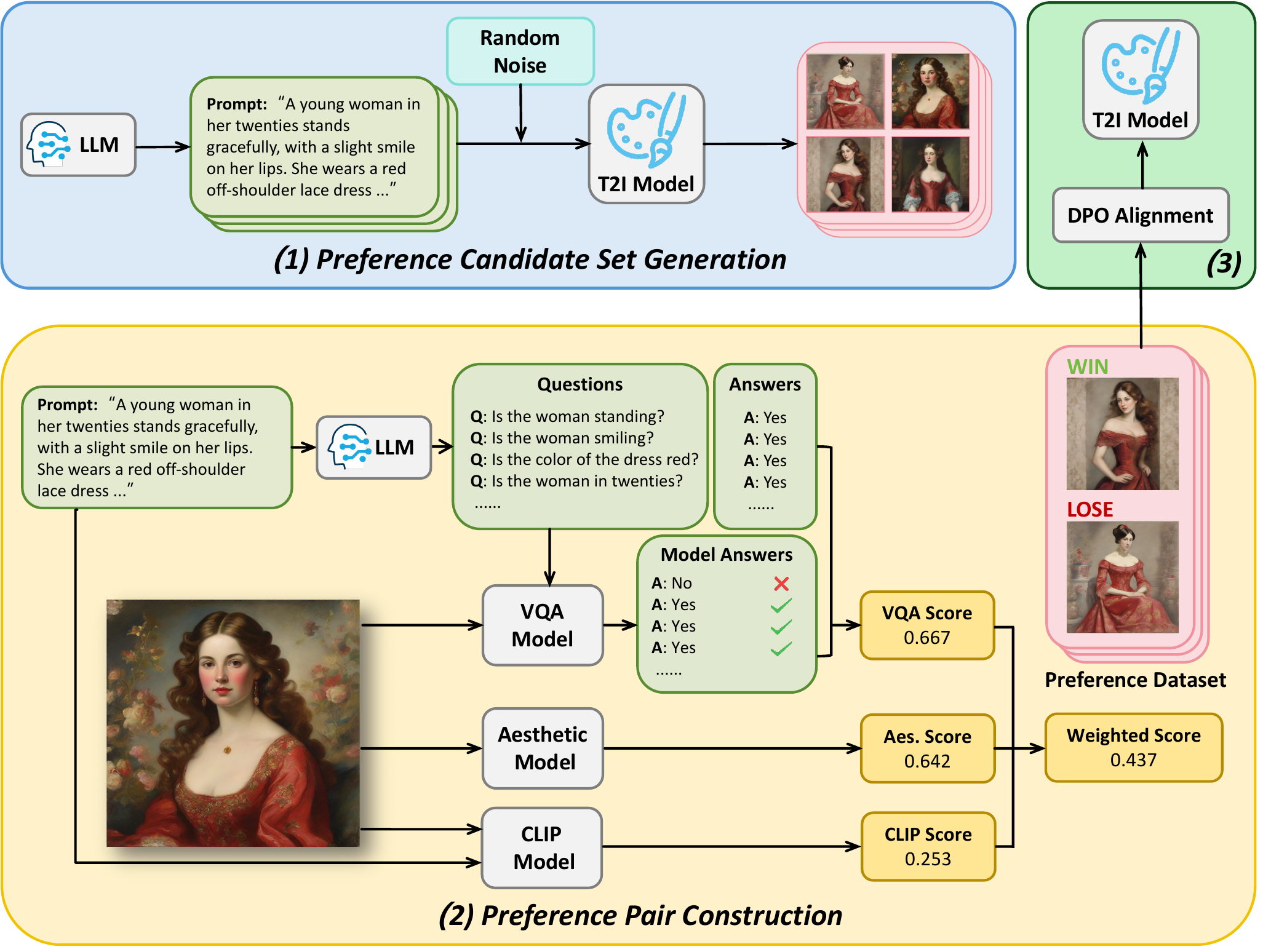}
  \caption{Overall pipeline of \modelname{}, which mainly encompasses 3 steps. \modelname{} learns from AI-generated feedback data with DPO. \modelname{} requires no human annotation, model architecture changes, or reinforcement learning.}
  \label{fig:pipeline}
\end{figure*}

The overall pipeline of \modelname{} is illustrated in \cref{fig:pipeline}.

\subsection{Preference Candidate Set Generation}

To encourage the diffusion model to generate diverse style images for further text-image pair preference datasets, we employ LLM to construct prompts that serve as image captions \(\bm{c}\).

We employ LLM to generate image captions \(\bm{c}\) from the instruction that would further feed into the T2I diffusion model. We encourage the LLM to generate 12 distinct categories for diversity: Natural Landscapes, Cities and Architecture, People, Animals, Plants, Food and Beverages, Sports and Fitness, Art and Culture, Technology and Industry, Everyday Objects, Transportation, and Abstract and Conceptual Art.

For each category, we utilize in-context learning strategy -- carefully craft 5 high-quality examples aimed at guiding the large language model to grasp the core characteristics and contexts of each category, thereby generating new prompts with relevant themes and rich content. Additionally, we emphasize the diversity in prompt lengths, aiming to produce both succinct and elaborate prompts to cater to different generational needs and usage scenarios.

To construct the preference candidate set, we consider a text-conditioned generative diffusion model \( G \) for candidate the image generation, where \( G \) accept input parameters: text condition \(\bm{c}\) and latent space noise \(\bm{z}_0\). We let the diffusion model to generate \(N\) candidate images. To enhance the diversity and distinctiveness of the images produced by the model, we incorporate Gaussian noise into the conditional input $\bm{c}$ and generate $\bm{z}_0$ with different random seeds. This approach aims to introduce more randomness and variation to avoid overly uniform or similar generated images. Specifically, the process of generating backup images can be represented as in \cref{eq:add_noise_in_diffusion}:
\begin{equation}
\bm{x}_0 = \text{G}(\bm{c} + \bm{n}, \bm{z}_0)
\label{eq:add_noise_in_diffusion}
\end{equation}
where Gaussian noise $\bm{n} \sim \mathcal{N}(0, \sigma^2\bm{I})$ is added to the conditional input, increasing the diversity of images.

In practice, by adjusting the value of variance $\sigma$ and using different random seeds to generate $\bm{z}_0$, the diversity of the generated images can be controlled. A larger $\sigma$ value will lead to greater variability in the conditility of the input, but potentially producing more diverse images but might also decrease the relevance of the image to the condition.

Therefore, we currently have the sample \(\bm{c}\) and its corresponding \(N\) preference candidate generated images. Next, we will filter and refine these candidates to construct the final preference pair dataset.

\subsection{Preference Pair Construction}

\subsubsection{VQA Questions Generation}
We also employ the LLM to refine the prompts generated for T2I generation into a series of question-and-answer pairs (QA pairs). By letting Visual Question Answering (VQA) model to answer these questions based on the generated images, the VQA score is calculated. We will establish the preference pair according to multiple image quality scores later.

To make the score easier to calculate, we ensure that the answers to these questions are uniformly ``yes'' in the instruction prompt. To refine the questions, we let the LLM to validate the questions if they are ambiguous or unrelated to the captions, therefore all questions are generated not valid or closely related to the text for answering by the validation process in the instruction prompt.

\subsubsection{VQA Score} The VQA score is computed by evaluating the correctness of answers provided by the VQA model to the questions generated from the text prompt \(\bm{c}\). For each text prompt \(\bm{c}\), the set of QA pairs is denoted as \(\{ (Q_i (\bm{c}), A_i (\bm{c})) \}\) for \(i=1, \dots, N_{\bm{c}}\), where \(N_{\bm{c}}\) is the total number of QA pairs generated for the text prompt \(\bm{c}\), and \(\bm{x}_0\) represents the image generated from \(\bm{c}\).

The VQA model \(\Phi\) is employed to answer all questions \({Q_i (\bm{c})}_{i=1}^{N_{\bm{c}}}\) based on the image \(\bm{x}_0\). The correctness of the VQA model's answers is evaluated by comparing them to the correct answers \(A_i (\bm{c})\). The VQA score~\cite{hu2023tifa}, which quantifies the consistency between the text and the generated image, is calculated in \cref{eq:vqa_score}:
\begin{equation}
\label{eq:vqa_score}
  s_\text{VQA} = \frac{1}{N_{\bm{c}}} \sum_{i=1}^{N_{\bm{c}}} 
  \begin{cases} 
  1 & \text{if } \Phi(\bm{x}_0, Q_i (\bm{c})) = A_i (\bm{c}), \\
  0 & \text{otherwise}.
  \end{cases}
\end{equation}
Here, the case structure explicitly represents the indicator function, which is \(1\) if the VQA model's answer matches the correct answer \(A_i (\bm{c})\), and \(0\) otherwise.

\subsubsection{CLIP Score}
Utilizing the CLIP~\cite{radford2021clip} model, we convert the prompt words and the generated image into vector representations, denoted as $\bm{c}^{(emb)}$ for text and $\bm{x}'_0$ for the image. The cosine similarity between the two vectors, computed in a shared embedding space, quantifies the alignment between the text and the image, embodying the CLIP Score~\cite{hessel2021clipscore}, defined in \cref{eq:clip_score}:
\begin{equation}
\label{eq:clip_score}
s_\text{CLIP} = \cos(\bm{c}^{(emb)}, \bm{x}'_0) = \left( \frac{\bm{c}^{(emb)}}{||\bm{c}^{(emb)}||_2} \cdot \frac{\bm{x}'_0}{||\bm{x}'_0||_2} \right) * \gamma
\end{equation}

\subsubsection{Aesthetic Score}
The aesthetic score assesses an image's visual appeal by analyzing multifaceted elements like composition, color harmony, style, and high-level semantics, which collectively contribute to the aesthetic quality of an image~\cite{ke2023Aesthetics}. The evaluation is defined in \cref{eq:aesthetic_score}:
\begin{equation}
\label{eq:aesthetic_score}
s_\text{Aesthetic} = AestheticModel(\bm{x}_0)
\end{equation}
where $\bm{x}_0$ signifies the input image, and $AestheticModel(\cdot)$ refers to a sophisticated model function that yields a score reflecting the image's aesthetic appeal on a normalized scale. Higher scores denote a greater aesthetic appeal.

\subsubsection{Weighted Score Calculation}
Consider a set of scores $\{s_1, s_2, \dots, s_n\}$, where each score $s_i$ corresponds to a distinct evaluation metric utilized. Alongside these scores, let there be a set of weights $W = \{w_1, w_2, \dots, w_n\}$, with each weight $w_i$ specifically assigned to modulate the influence of its corresponding score $s_i$.

The composite score for an image $\bm{x}_0$, which integrates these diverse evaluation metrics, is determined by calculating the sum of the weighted scores. The formula for computing this aggregated score is given by \cref{eq:weighted_score}:
\begin{equation}
\label{eq:weighted_score}
S(\bm{x}_0) = \sum_{i=1}^{n} w_i s_i (\bm{x}_0)
\end{equation}
where $n$ represents the total number of individual scores. The weighted sum approach facilitates the model's capability to assess images across varied criteria, offering a comprehensive understanding of the image's quality and relevance.

\subsubsection{Preference Pair Dataset Construction} With the generated set of $N$ images $\bm{X}_0 = \{\bm{x}^1_0, \bm{x}^2_0, \dots, \bm{x}^N_0\}$ for a given textual prompt $\bm{c}$, each candidate image is then evaluated to assign the score calculated in multiple aspects as the aformentioned weighted score. To identify the most and least preferred images, which termed as the ``winner'' and ``loser'', we apply the selection criteria in \cref{eq:win_sel} and \cref{eq:lose_sel}:
\begin{equation}
\label{eq:win_sel}
\bm{x}^w_0 = \arg\max_{\bm{x}^i_0 \in \bm{X}_0} S(\bm{x}^i_0)
\end{equation}
\begin{equation}
\label{eq:lose_sel}
\bm{x}^l_0 = \arg\min_{\bm{x}^i_0 \in \bm{X}_0} S(\bm{x}^i_0)
\end{equation}
This approach yields a preference pair for each textual prompt \(\bm{c}\), represented as $(\bm{c}, \bm{x}^{w}_0, \bm{x}^{l}_0)$. The rationale behind selecting the highest and lowest scored images is to capture the widest possible discrepancy in quality and relevance, providing a clear contrast suitable for finetuning with DPO.

\subsection{DPO Alignment}

Derive from Diffusion-DPO~\cite{rafailov2023DPO}, we consider the preference dataset, denoted as $\mathcal{D} = \{(\bm{c}, \bm{x}^{w}_0, \bm{x}^{l}_0)\}$. Applying DPO for diffusion models is modeled as the following objective function \(L(\theta)\) in \cref{eq:diffusion_dpo}. For the detailed notation of algorithms \cref{eq:diffusion_dpo}, please refer to Diffusion-DPO~\cite{rafailov2023DPO} and DPO~\cite{rafailov2023DPO}.
\begin{equation}
\label{eq:diffusion_dpo}
\begin{aligned}
&L(\theta) = -\mathbb{E}_{(\bm{x}_0^w,\bm{x}_0^l) \sim \mathcal{D}, t \sim \mathcal{U}(0,T), \bm{x}_t^w \sim q(\bm{x}_t^w|\bm{x}_0^w), \bm{x}_t^l \sim q(\bm{x}_t^l|\bm{x}_0^l)} \\
&\quad \quad \quad \log \sigma(-\beta T \omega(\lambda_t)(\|\boldsymbol{\epsilon}^w - \boldsymbol{\epsilon}_{\theta} (\bm{x}^w_t, t, \bm{c})\|_2^2 \\
&\quad \quad \quad- \|\boldsymbol{\epsilon}^w - \boldsymbol{\epsilon}_{\text{ref}} (\bm{x}^w_t, t, \bm{c})\|_2^2 - (\|\boldsymbol{\epsilon}^l - \boldsymbol{\epsilon}_{\theta} (\bm{x}^l_t, t, \bm{c})\|_2^2 \\
&\quad \quad\quad - \|\boldsymbol{\epsilon}^l - \boldsymbol{\epsilon}_{\text{ref}} (\bm{x}^l_t, t, \bm{c})\|_2^2)))
\end{aligned}
\end{equation}
where $\bm{x}^*_t = \alpha_t \bm{x}^*_0 + \sigma_t \boldsymbol{\epsilon}^*$, $\boldsymbol{\epsilon}^* \sim \mathcal{N}(0,\bm{I})$. Here, $\alpha_t$ and $\sigma_t$ are the noise scheduling functions as defined in~\cite{rombach2022highresolution}. Consequently, $\bm{x}_t \sim q(\bm{x}_t|\bm{x}_0) = \mathcal{N}(\bm{x}_t; \alpha_t \bm{x}_0, \sigma_t^2 \bm{I})$. Similar to~\cite{wallace2023diffusionDPO}, we incorporate $T$ and $\omega(\lambda_t)$ into the constant $\beta$.

\section{Experimental Setups}

\subsection{Datasets}

To evaluate whether our \modelname{} can enhance the performance of text-to-image models across a wide range of prompts, we consider the following benchmarks: 
\begin{enumerate}
  \item\textbf{TIFA}~\cite{hu2023tifa}: Based on the correct answers to a series of predefined questions. TIFA employs visual question answering (VQA) models to determine whether the content of generated images accurately reflects the details of the input text. The benchmark itself is comprehensive, encompassing 4,000 different text prompts and 25,000 questions across 12 distinct categories. 
  \item\textbf{HPS v2}~\cite{wu2023human}: Human Preference Score v2 (HPS v2) is a benchmark designed to evaluate models' capabilities across a variety of image types. It comprises 3,200 distinct image captions and covers five categories of image descriptions: anime, photo, drawbench, concept-art, and paintings.
\end{enumerate}

\subsection{Hyperparameters}

For each given text prompt $\bm{c}$, we let the diffusion model generate \(N=8\) samples as backup images for preference dataset construction. In this process, we add Gaussian noise \(\bm{n} \sim \mathcal{N}(0, \sigma^2\bm{I})\) to the text embedding, where \(\sigma\) is set to \(0.1\). In the calculation of CLIP score, \(\gamma\) is set to $100$, which leads to the CLIP Score range between 0 and 100. We also rescale the VQA score and aesthetic score to $0-100$ by multiplying the original score by $100$.

The weighting of each score measurements is allocated as: \(w_{\text{VQA}}=0.35\), \(w_{\text{CLIP}}=0.55\), \(w_{\text{Aesthetic}}=0.1\). Thus, the weighted score $S$ for an image $\bm{x}$ is calculated as \cref{eq:weighted_score_weights}:
\begin{equation}
\label{eq:weighted_score_weights}
\begin{aligned}
S(x) = 0.35 s_{\text{VQA}}(x) + 0.55 s_{\text{CLIP}}(x) + 0.1 s_{\text{Aes.}}(x)
\end{aligned}
\end{equation}

During the DPO alignment stage, we finetune the original diffusion model. For the SD v1.4 and SD v1.5 models, the learning rate is 5e-7, the batch size is 128, the output image size is $512 \times 512$. For the SDXL-base model, the learning rate is 1e-6, the batch size is 64, the output image size is $1024 \times 1024$. We finetune the diffusion model for 1,000 steps. The random seed is set to 200 in \cref{fig:aesthetic_images} and \cref{fig:faithful_images}.

\subsection{Baseline Models and Utilized Models}

We evaluate \modelname{} using Stable Diffusion v1.4 (SD v1.4), Stable Diffusion v1.5 (SD v1.5)~\cite{rombach2022sd15}, and Stable Diffusion XL Base 1.0 (SDXL-base)~\cite{podell2023sdxl}, widely acknowledged in related research as the current leading open-source text-to-image (T2I) models. For prompt construction, we employ ChatGPT (GPT-3.5)~\cite{blog2023ChatGPT}. For generating Q\&A pairs, we use Gemini Pro~\cite{pichai2023gemini}. Both are accessed through their official API. In addition, we adopt \textit{Salesforce/blip2-flan-t5-xxl} for VQA scoring model~\cite{li2023blip2}, \textit{openai/clip-vit-base-patch16} for evaluating CLIP score~\cite{hessel2021clipscore}, \textit{Vila}~\cite{ke2023Aesthetics} for calculating aesthetic score~\cite{ke2023Aesthetics}, which are consistent with the baseline methods' settings. We also employ GPT-4 Vision (GPT-4V)~\cite{openai2023gpt4} to simulate human preferences when evaluating the image quality.

\section{Experimental Results}

\subsection{Benchmarking Results on HSP v2}

\subsubsection{Evaluate by CLIP Score and Aesthetic Score}

\begin{figure}[!ht]
  \centering
  \begin{subfigure}[b]{0.48\linewidth}
    \centering
    \includegraphics[width=1.0\linewidth]{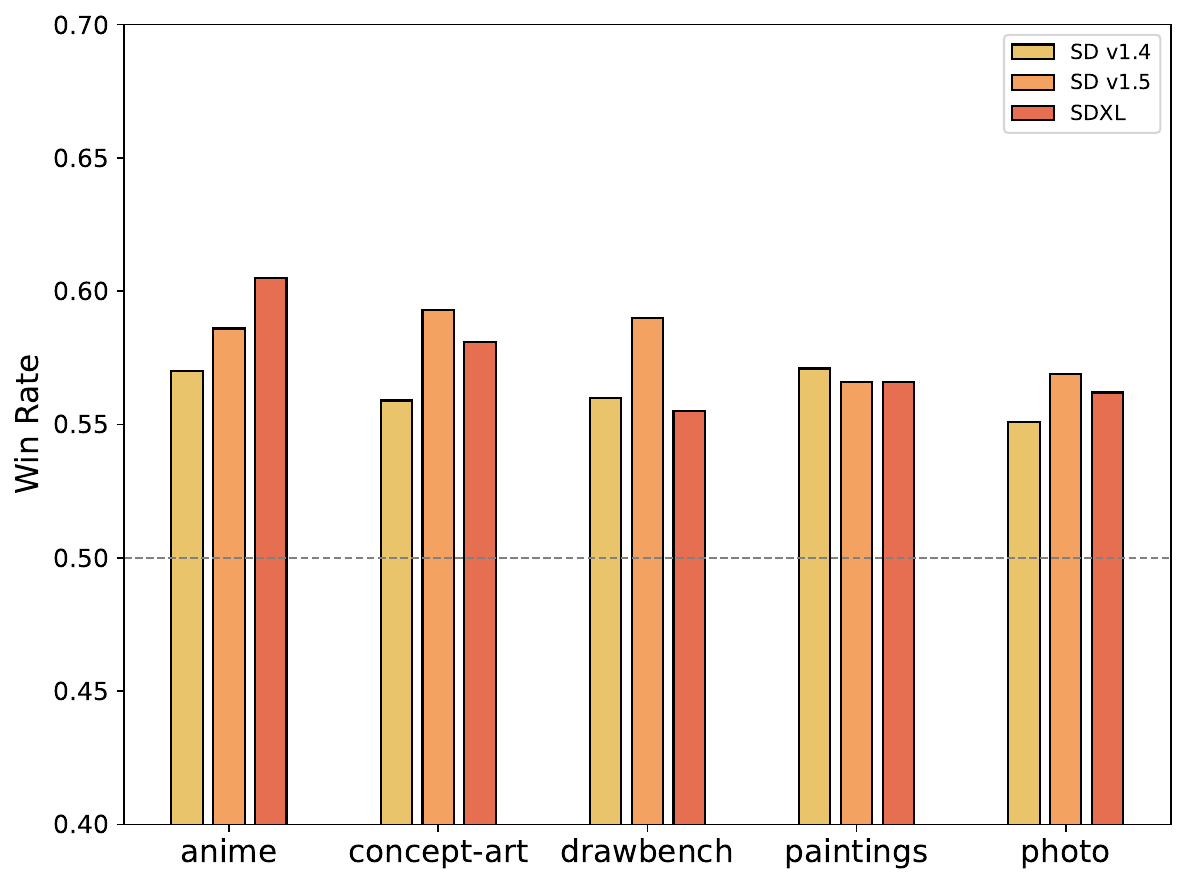}
    \caption{CLIP score win rates}
    \label{fig:clip_score}
  \end{subfigure}
  \hfill 
  \begin{subfigure}[b]{0.48\linewidth}
    \centering
    \includegraphics[width=1.0\linewidth]{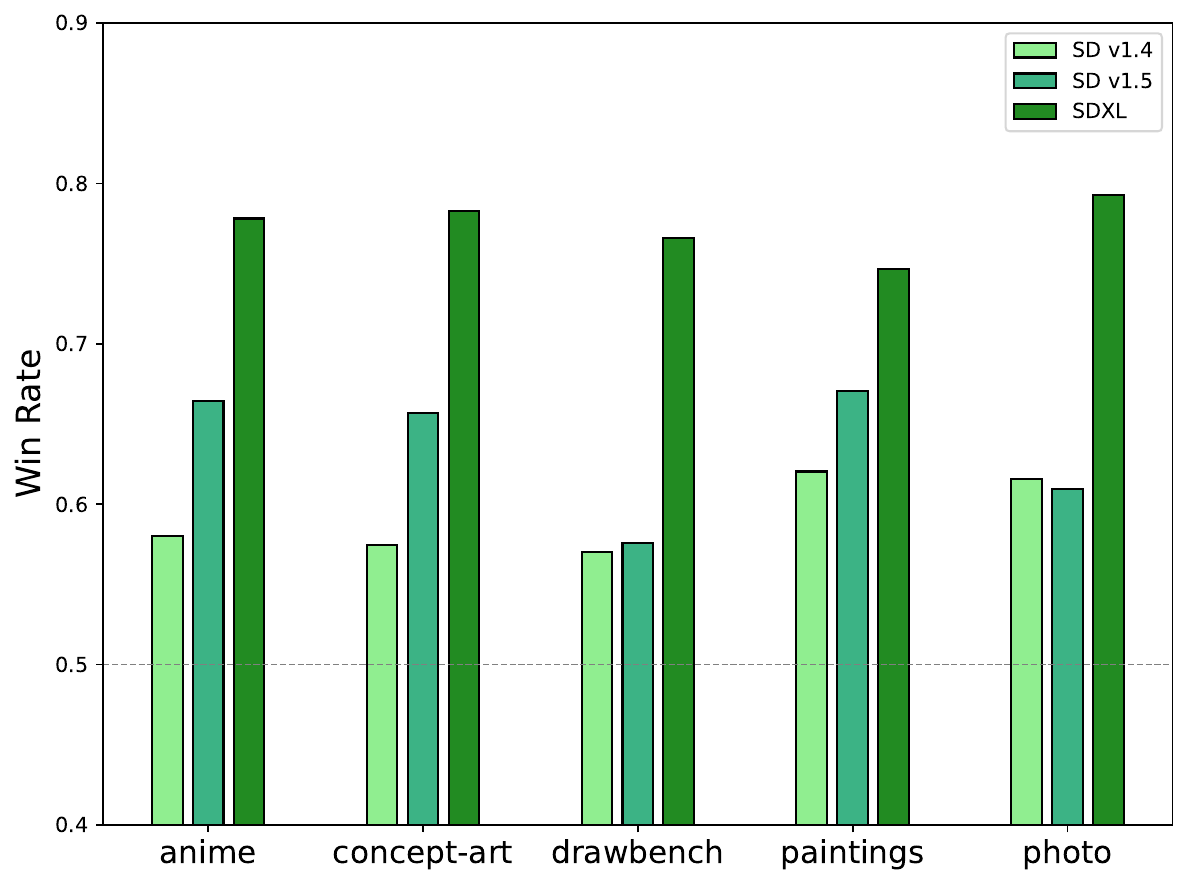}
    \caption{Aesthetic score win rates}
    \label{fig:saes_score}
  \end{subfigure}
  \caption{Comparison of the win rates of SD v1.4, SD v1.5 and SDXL-base with or without our \modelname{} on HPS v2. CLIP score (left) and aesthetic score (right).}
  \label{fig:clip_aes_comparison}
  \vspace{-3mm}
\end{figure}

As in \cref{fig:clip_aes_comparison}, we test the win rates of models finetuned with our method \modelname{} against the original models on the CLIP score and aesthetic score in the HSP v2 benchmark. The experimental results show that with \modelname{}, models consistently achieve win rates exceeding 50\% across all image categories and both evaluation metrics,  CLIP score and aesthetic score, compared to the original baseline SD v1.4, SD v1.5, and SDXL-base models without finetuning. Notably, after \modelname{} finetuning, the SDXL-base model not only achieves a win rate of 60.5\% in the CLIP score compared to the original model in the anime category images, but also achieves a win rate of 77.4\% in the aesthetic score for the same category images. The average win rate of the CLIP score and aesthetic score for the three models increases to 57.2\% and 61.6\%, respectively, compared to base ones.

\subsubsection{Evaluate by GPT-4 Vision to Simulate Human Preference}

In this study, we explore the efficacy of \modelname{} in enhancing image generation models, leveraging the capabilities of GPT-4 Vision (GPT-4V) as reported by OpenAI in 2023~\cite{openai2023gpt4} to simulate human preferences. Our methodology involves: (1) a comparative analysis of images generated by various diffusion models before and after the application of \modelname{}; and (2) a comparative analysis of images generated by the SD v1.4 model after applying \modelname{} or other alignment methods. These images, accompanied by their respective descriptions, are submitted to GPT-4V for evaluation based on three critical aspects: \textbf{General Preference (Q1)}: ``Which image do you prefer?''; \textbf{Prompt Alignment (Q2)}: ``Which image better fits the text description?''; \textbf{Visual Appeal (Q3)}: ``Disregarding the prompt, which image is more visually appealing?''.

\begin{table}[!ht]
\caption{Win rate results of using GPT-4V to evaluate our finetuned models based on SD v1.4, SD v1.5, and SDXL-base, compared to the original models and models aligned with DDPO, Structured Diffusion, and SynGen (only on SD v1.4), for general preference (Q1), prompt alignment (Q2), and visual appeal (Q3) on the HSP v2 dataset.}
\label{tab:gpt4v_evaluation}
\resizebox{\linewidth}{!}{
\begin{tabular}{c|c|ccc}
\toprule
\textbf{Test Model}               & \textbf{Method}                     & \textbf{General} & \textbf{Faithful} & \textbf{Aesthetic} \\ \midrule
\multirow{4}{*}{SD v1.4} & vs Original                                                      & 62\%    & 58\%     & 65\%      \\
                         & vs DDPO                                                          & 68\%    & 78\%     & 82\%      \\
                         & vs Structured Diffusion & 64\%    & 70\%     & 79\%      \\
                         & vs SynGen                                                        & 61\%    & 58\%     & 58\%      \\ \midrule
SD v1.5                  & vs Original                                                      & 68\%    & 67\%     & 65\%      \\ \midrule
SDXL-base                     & vs Original                                                      & 62\%    & 69\%     & 76\%      \\
\bottomrule
\end{tabular}
}
\end{table}

The evaluation process involves collecting and analyzing the frequency with which images produced by both the original and the finetuned model are favored under each question category. The results of the comparative analysis of images generated by various diffusion models before and after the application of \modelname{} are presented in \cref{tab:gpt4v_evaluation}, which sequentially displays the performance metrics. The performance reveals that adding \modelname{} yields substantial enhancements across all models concerning Q1, Q2, and Q3. Notably, with our \modelname{} applied, we achieve an average of 62\%, 67\%, and 69\% win rates across the three aspects for the SD v1.4, SD v1.5, and SDXL-base models respectively. The results of the comparative analysis of images generated by the SD v1.4 model after applying \modelname{} or other alignment methods are also presented in \cref{tab:gpt4v_evaluation}. The performance shows \modelname{} consistently outperforms other baselines (DDPO, Structured Diffusion, and SynGen) applied to the SD v1.4 in all dimensions, achieving average win rates of 64.2\%, 68.4\%, and 72.8\% respectively, when evaluated by GPT-4V on images generated from the HPS v2. These results demonstrate the effectiveness of \modelname{} in enhancing performance under various prompts.

\subsubsection{Human Evaluation of GPT-4V Judgments}
To validate the efficacy of GPT-4V for image evaluation and address potential biases in AI assessments, we compare its consistency with human evaluations. We randomly select 9 pairs of images generated by the \modelname{} that are favored by GPT-4V. A total of 58 graduate students from China participate in the evaluation. Each participant assesses each image pair based on the criteria Q1, Q2, Q3. Each image pair is independently rated on these three dimensions, resulting in 27 questions per participant (9 image pairs $\times$ 3 dimensions). For Q1, there is 78\% agreement between GPT-4V and human evaluations, for Q2, 83\%, and for Q3, 70\%. All dimensions show agreement rates above 50\%, indicating that GPT-4V's evaluations align closely with human preferences and confirming its reliability as a tool for reducing individual biases and maintaining objectivity in image evaluation.

\subsection{Benchmarking Results on TIFA}


In \cref{tab:tifa_vqa_score}, we further test our method on the TIFA benchmark, highlighting \modelname{}'s SOTA performance on VQA score and aesthetic score over other latest SOTA alignment methods. Specifically, we compare three types of alignment methods: training-free approaches capable of modifying outputs without retraining the model, such as StructureDiffusion~\cite{feng2023trainingfree} and SynGen~\cite{rassin2024linguistic}; reinforcement learning (RL)-based methods aimed at improving model outputs, such as DPOK~\cite{fan2024reinforcement} and DDPO~\cite{black2023trainingDDPO}; and methods like DreamSync~\cite{sun2023dreamsync}, which employ self-training strategy but focus on SFT stage. Given that these baseline methods are all based on SD v1.4, we ensure a fair comparison by using the same version of the SD model as the foundation and employing the same VQA model (BLIP-2) for evaluation. Results reveal that our method \modelname{} can simultaneously improve the text fidelity and visual quality of SD v1.4, SD v1.5, and SDXL-base models. For SD v1.4, \modelname{} achieves an improvement of 1.3\% of VQA score and 3.3\% of aesthetic score, with a total improvement of 4.6\% on the TIFA benchmark, higher than all baseline models. Note that although DPOK shows a 1.9\% improvement on aesthetic score, it reduces the model's text faithfulness through VQA score. For SD v1.5 and SDXL-base, our method \modelname{} leads to improvements of 1.6\% and 1.1\% for SD v1.5, 1.3\% and 4.3\% for SDXL-base on VQA score and aesthetic score respectively, which are both higher than the results achieved by DreamSync finetuned using self-training SFT.

\begin{table}[!ht]
\vspace{-2mm}
\centering
\caption{Results of different alignment methods on VQA score and aesthetic score on the TIFA benchmark. Red indicates improvement, while Green indicates a decrease. The best scores for each model type are in Bold. Column ``Sum'' denotes the sum of improvements on \(s_{\text{VQA}}\) and \(s_{\text{Aes.}}\)}
\label{tab:tifa_vqa_score}
\vspace{-1mm}
\resizebox{\linewidth}{!}{
\begin{tabular}{lllllc}
\toprule
\multicolumn{1}{c}{\textbf{Model}} &    & \multicolumn{1}{c}{\textbf{Alignment}}          & \multicolumn{1}{c}{\textbf{\(s_{\text{VQA}}\)}} & \multicolumn{1}{c}{\textbf{\(s_{\text{Aes.}}\)}} & \textbf{Sum} \\
\midrule
\multirow{7}{*}{SD v1.4} &                                                    & No alignment       & 76.6                          & 44.6 & -                                \\ \cline{2-3}
                         & \multicolumn{1}{c}{\multirow{2}{*}{Training-Free}} & SynGen             & 76.8 \textcolor{red}{($+$0.2)}                    & 42.4 \textcolor{green!60!black}{($-$2.2)}  &$-$2.0                        \\
                         & \multicolumn{1}{c}{}                               & StructureDiffusion & 76.5 \textcolor{green!60!black}{($-$0.1)}         & 41.5 \textcolor{green!60!black}{($-3.1$)}  &$-$3.0                        \\ \cline{2-3}
                         & \multicolumn{1}{c}{\multirow{2}{*}{RL}}            & DPOK               & 76.4 \textcolor{green!60!black}{($-$0.2)}         & 46.5 \textcolor{red}{($+$1.9)}    &$+$1.7                      \\
                         & \multicolumn{1}{c}{}                               & DDPO               & 76.7 \textcolor{red}{($+$0.1)}                    & 43.5 \textcolor{green!60!black}{($-$1.1)}  &$-$1.0                        \\ \cline{2-3}
                         & \multicolumn{1}{c}{\multirow{2}{*}{Self-Training}} & DreamSync          & 77.6 \textcolor{red}{($+$1.0)}                    & 44.9 \textcolor{red}{($+$0.3)}             &$+$1.3             \\
                         & \multicolumn{1}{c}{}                               & \modelname{} (Ours)    & \textbf{77.9} \textcolor{red}{($+$1.3)}         & \textbf{47.9} \textcolor{red}{($+$3.3)}           &\textbf{$+$4.6}               \\
\midrule
\multirow{3}{*}{SD v1.5} & \multirow{3}{*}{}                   & No alignment       & 77.1                          & 48.0   &-                             \\
&         & DreamSync      &         77.7 \textcolor{red}{($+$0.6)}                 &  47.6 \textcolor{green!60!black}{($-$0.4)} &    $+$0.2                           \\
                         &                                                    & \modelname{} (Ours)    & \textbf{78.7} \textcolor{red}{($+$1.6)}         & \textbf{49.1} \textcolor{red}{($+$1.1)}  & \textbf{$+$2.7}                        \\
\midrule
\multirow{3}{*}{SDXL-base}    & \multirow{3}{*}{}                                  & No alignment       & 82.0                          & 60.9  &-                              \\
              &         & DreamSync      &       83.1 \textcolor{red}{($+$1.1)}                    & 64.1 \textcolor{red}{($+$3.2)}  &$+$4.3                            \\
                         &                                                    & \modelname{} (Ours)    & \textbf{83.3} \textcolor{red}{($+$1.3)}         & \textbf{65.2} \textcolor{red}{($+$4.3)} & \textbf{$+$5.5}                         \\
\bottomrule
\end{tabular}
}
\vspace{-4mm}
\end{table}

\subsection{Experiment of Comparing the Dataset Quality between MJHQ-30K and \modelname{}}

MJHQ-30K is a benchmark dataset used for automatically evaluating the aesthetic quality of models~\cite{li2024playground}. It consists of high-quality images curated from Midjourney, covering 10 common categories, with each category containing 3,000 samples.
MJHQ-30K can also serve as a training dataset for general SFT. To compare the quality of the preference dataset built using \modelname{} with MJHQ-30K, we finetune SD v1.4, SD v1.5 and SDXL-base using MJHQ-30K and compare their performance against the SD v1.4, SD v1.5 and SDXL-base finetuned with \modelname{}. As shown in \cref{tab:vs_MJHQ30K}, \modelname{} applied to SD v1.4, SD v1.5 and SDXL-base achieve superior improvements in text alignment. Although finetuning SD v1.4 and SD v1.5 with the MJHQ-30K dataset results in the highest improvement in aesthetic scores, this is because the images in MJHQ-30K is generated by Midjourney, which have much higher aesthetic quality than those generated by SD v1.4 and SD v1.5 for self-training. When finetuning SDXL-base with MJHQ-30K, the improvement in aesthetic scores is less pronounced compared to \modelname{}, demonstrating the effectiveness of the preference dataset constructed using \modelname{}.

\begin{table}[!ht]
\caption{SD v1.4, SD v1.5 and SDXL-base's results of general SFT setting on MJHQ-30K compared to \modelname{}.}
\label{tab:vs_MJHQ30K}
\footnotesize
\centering
\resizebox{\linewidth}{!}{
\begin{tabular}{llllc}
\toprule
\multicolumn{1}{c}{\textbf{Model}} & \multicolumn{1}{c}{\textbf{Alignment}} & \multicolumn{1}{c}{\textbf{\(s_{\text{VQA}}\)}} & \multicolumn{1}{c}{\textbf{\(s_{\text{Aes.}}\)}} & \multicolumn{1}{c}{\textbf{Sum}}\\ \hline
\multirow{3}{*}{SD v1.4}           
& No alignment                                & 76.6                              & 44.6                & -            \\ \cline{2-5}
                                    & MJHQ-30K+SFT                                & 77.6 \textcolor{red}{(+1.0)}                             & \textbf{48.3} \textcolor{red}{(+3.7)}        & \textbf{+4.7}                    \\ \cline{2-5} 
                                    & \modelname{} (Ours)                           & \textbf{77.9} \textcolor{red}{(+1.3)}                             & 47.9 \textcolor{red}{(+3.3)}     & +4.6                       \\ \hline
\multirow{3}{*}{SD v1.5}            & No alignment                                & 77.1                              & 48.0  & -                           \\ \cline{2-5} 
                                    & MJHQ-30K+SFT                                & 78.3 \textcolor{red}{(+1.2)}                             & \textbf{49.3} \textcolor{red}{(+1.3)}   & +2.5                         \\ \cline{2-5} 
                                    & \modelname{} (Ours)                           & \textbf{78.7} \textcolor{red}{(+1.6)}                             & 49.1 \textcolor{red}{(+1.1)}   & \textbf{+2.7}                         \\ \hline
\multirow{3}{*}{SDXL-base}            & No alignment                                & 82.0                              & 60.9  & - \\ \cline{2-5}  & MJHQ-30K+SFT                                &          82.6 \textcolor{red}{(+0.6)}                        &           61.1 \textcolor{red}{(+0.2)}   & +0.8\\ \cline{2-5}
                                    & \modelname{} (Ours)                           & \textbf{83.3} \textcolor{red}{(+1.3)}                             & \textbf{65.2} \textcolor{red}{(+4.3)}  & \textbf{+5.6}                          \\
\bottomrule
\end{tabular}
}
\vspace{-2mm}
\end{table}

\subsection{Experiment of Gaussian Noise for Diversity}

\begin{figure}[!ht]
  \centering
  \includegraphics[width=0.8\linewidth]{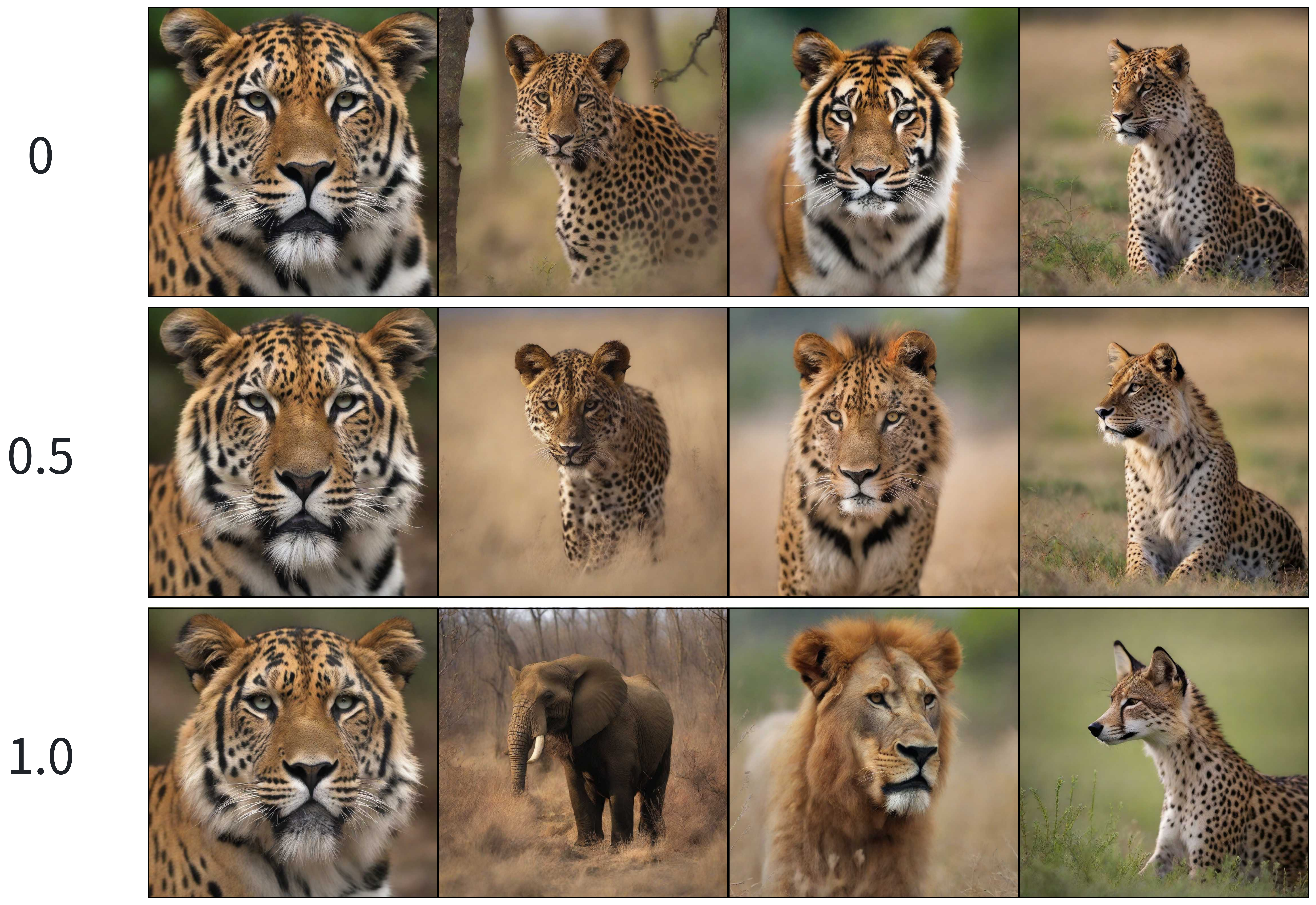}
  \caption{Impact of noise on image diversity. With the numbers on the left side of the images indicating the increasing weight of noise, four images were generated using the same text input ``wild animal''.}
  \label{fig:diversity}
  \vspace{-5mm}
\end{figure}

To demonstrate that adding Gaussian noise \(n\) to a given condition \(c\) during the generation of \(N\) candidate images significantly enhances the diversity of the candidates, we conduct an experiment as shown in the \cref{fig:diversity}. We utilize a consistent text input ``wild animal'' to generate four images and systematically adjust the weight of the noise \(n\). By comparing the generated images under different noise weights, we observe significant changes in the variety of animal species with the increasing weight of the noise (\cref{fig:diversity}). Notably, when the noise weight is set to 1, the images exhibit the most diverse range of animal species. This finding supports our hypothesis that the introduction of Gaussian noise effectively expands the coverage of the conditional input, thus increasing the exploration space of the model and significantly enhancing the diversity of the generated images.

\subsection{Ablation Experiment of Multi-Aspect Scoring}

As depicted in \cref{tab:ablation_experiments}, to validate the efficacy of the three scores that we employ for image quality assessment, we conduct a thorough ablation study. We train the SD v1.5 model on preference datasets constructed with different combinations of the three scores, along with PickScore~\cite{kirstain2024pick}. As in \modelname{}, with training model on preference datasets built using a combination of CLIP score, VQA score, and aesthetic score result in the greatest improvement across all three metrics. While other combinations often show a decrease in certain metrics rather than a consistent improvement on all metrics.

\begin{table}[!ht]
\footnotesize
    \centering
    \caption{Results of applied scoring measures. Experiments are conducted with SD v1.5 on TIFA.}
    \label{tab:ablation_experiments}
    \vspace{-1mm}
    \begin{tabular}{ccccccc}
        \toprule
        \multicolumn{4}{c}{\textbf{Applied Measures}} & \multirow{2}{*}{\textbf{\(s_{\text{CLIP}}\)}} & \multirow{2}{*}{\textbf{\(s_{\text{VQA}}\)}} & \multirow{2}{*}{\textbf{\(s_{\text{Aes.}}\)}} \\ \cline{1-4}
        +CLIP & +VQA & +Aes. & +Pick & & & \\
        \midrule
        - & -& -& -& 27.0 & 77.1 & 48.0 \\
        \checkmark & -& -& -& \textcolor{red}{27.2} & \textcolor{red}{77.7} & \textcolor{green!60!black}{47.3} \\
        - & \checkmark & -& -& \textcolor{red}{27.1} & \textcolor{red}{77.4} & \textcolor{green!60!black}{45.7} \\
        - & -& \checkmark & -& 27.0 & \textcolor{green!60!black}{76.8} & \textcolor{red}{48.6} \\
        \checkmark & \checkmark & -& -& \textcolor{red}{27.2} & \textcolor{red}{77.5} & \textcolor{green!60!black}{47.0} \\
        \checkmark & -& \checkmark & -& \textcolor{red}{27.2} & \textcolor{red}{77.8} & \textcolor{green!60!black}{47.2} \\
        -& \checkmark & \checkmark & -& \textcolor{red}{27.1} & \textcolor{red}{77.2} & \textcolor{red}{48.2} \\
        -& -& -& \checkmark & \textcolor{red}{27.1} & \textcolor{red}{78.0} & \textcolor{green!60!black}{47.8} \\
        \midrule
        \checkmark & \checkmark & \checkmark & -& \textbf{\textcolor{red}{27.3}} & \textbf{\textcolor{red}{78.7}} & \textbf{\textcolor{red}{49.1}} \\
        \bottomrule
    \end{tabular}
    \vspace{-1mm}
\end{table}

\subsection{Qualitative Comparison of Faithfulness and Coherence}



\begin{figure}[!ht]
  \centering
  \begin{subfigure}[b]{0.95\linewidth}
    \includegraphics[width=\linewidth]{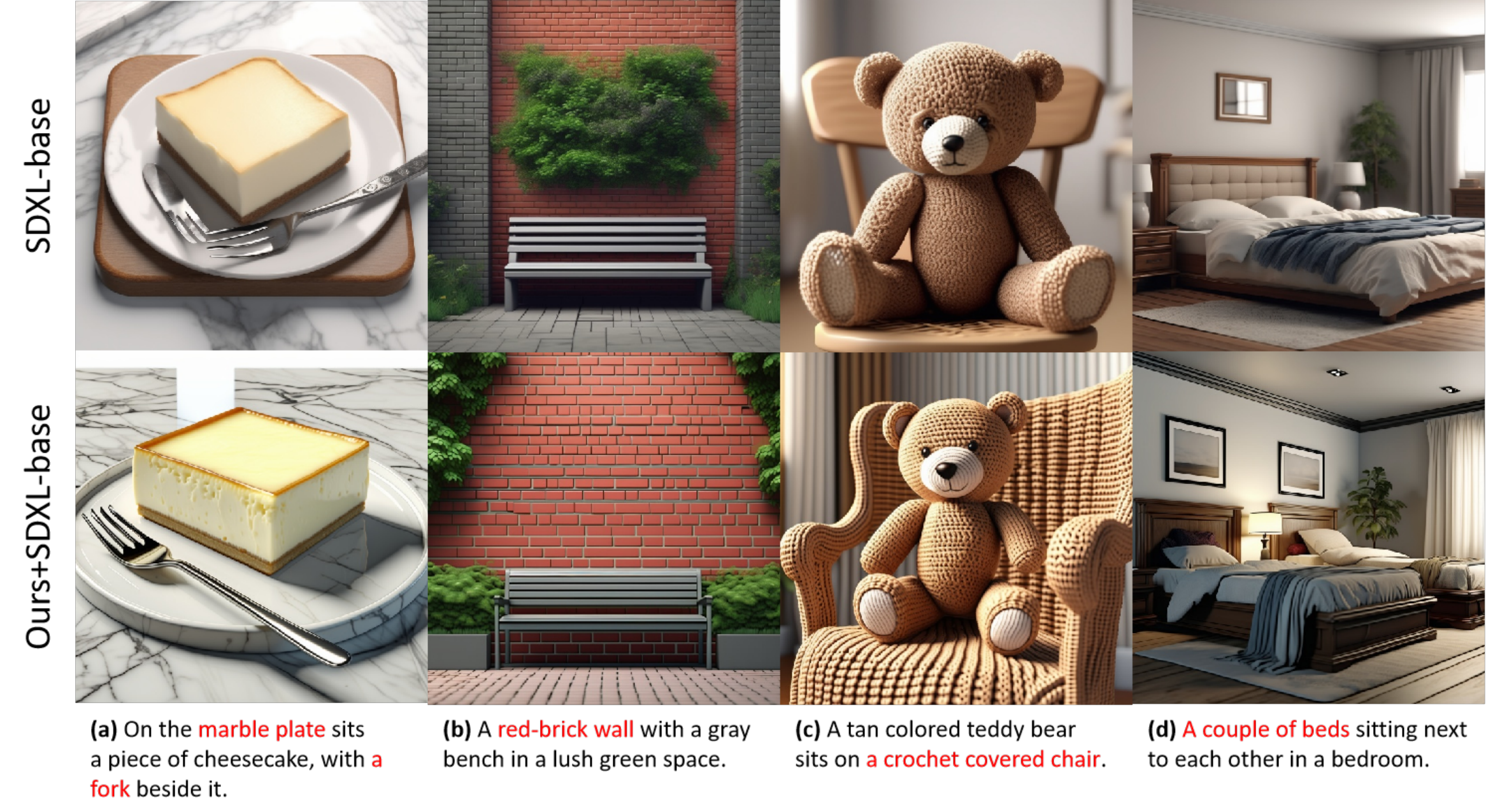}
    \caption{Text faithfulness comparison}
    \label{fig:faithful_images}
  \end{subfigure}
  \begin{subfigure}[b]{0.95\linewidth}
    \includegraphics[width=\linewidth]{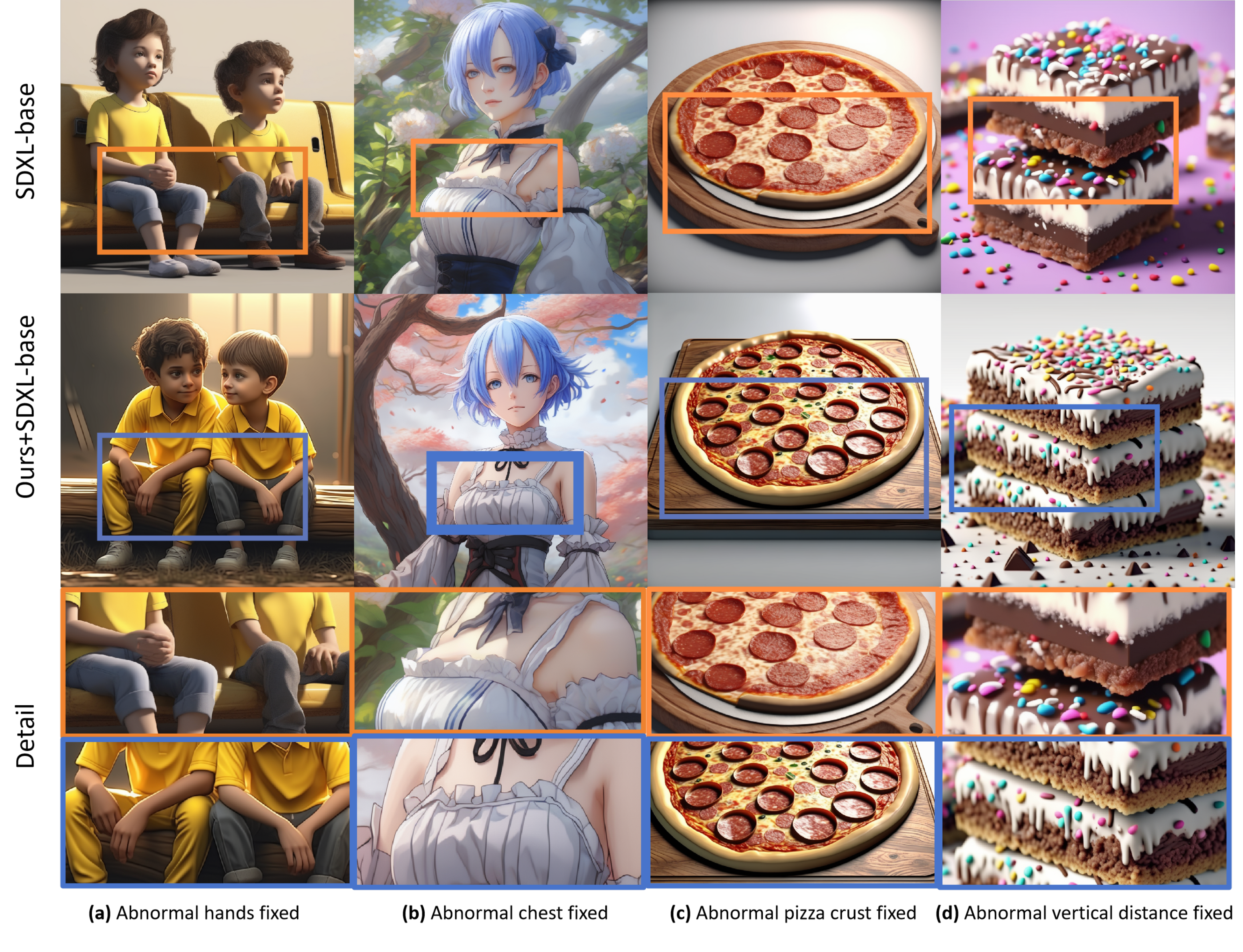}
    \caption{Adherence to real-world rules}
    \label{fig:aesthetic_images}
  \end{subfigure}
  \vspace{-3mm}
  \caption{Comparison between SDXL-base and \modelname{} (Ours)+SDXL-base. (a) Red-highlighted text indicates discrepancies with input prompts. (b) Third row compares details, showing \modelname{}'s improved coherence and detail.}
  \label{fig:combined_comparison}
  \vspace{-3mm}
\end{figure}

\cref{fig:faithful_images,fig:aesthetic_images} compare images generated by SDXL-base and \modelname{} using identical prompts. While SDXL-base generates vivid images, they sometimes deviate from input descriptions (\cref{fig:faithful_images}) or contain unrealistic details (\cref{fig:aesthetic_images}). For example, SDXL-base produces unnatural wrinkles on a girl's chest and physically impossible floating cakes. After finetuning with \modelname{}, the generated images show improved consistency with prompts and better adherence to real-world physics.


\vspace{-3mm}
\section{Conclusions}

This paper introduces a text-to-image generation framework \modelname{}. By leveraging Direct Preference Optimization (DPO) and multi-aspect AI feedback, \modelname{} significantly enhances the prompt following ability and image quality regarding style, coherence, and aesthetics. Extensive experiments on the HPSv2 and TIFA benchmark demonstrate that \modelname{} outperforms baseline models in terms of VQA scores, CLIP score, aesthetic evaluation. Based on an AI-driven feedback loop, \modelname{} eliminates the need for costly human-annotated data and manual intervention, paving the way for scalable alignment techniques.

\clearpage
\newpage

\section*{Acknowledgments}
This work was supported by the National Key R\&D Program of China (No. 2023YFB3309000), the National Natural Science Foundation of China under Grants U2241217, 62473027 and 62473029.
\bibliography{ref}

\appendix
\clearpage
\newpage
\onecolumn
\section{Notations}

The notations and their descriptions in the paper are shown in \cref{tab:notation}.

\begin{table}[!ht]
    \centering
    \caption{Notations symbols and their descriptions.}
    \label{tab:notation}
    \begin{tabular}{c|c}
\toprule
\textbf{Notations} & \textbf{Descriptions} \\
\midrule
\( \bm{c}, \bm{c}^{(emb)} \) & LLM-generated text prompt and its embedding \\
\( \text{G} \) & T2I diffusion model \\
\( \bm{n} \) & Gaussian noise \\
\( \bm{z}_0 \) & latent space noise in diffusion model \\
\(\bm{x}_{t}^{i}\) & the \(i\)-th diffusion model generated image of step \(t\) \\
\((Q_i (\bm{c}), A_i (\bm{c}))\) & \(i\)-th question-answer pair for text prompt \(\bm{c}\) \\
\( N_{\bm{c}} \) & the number of question-answer pairs \\
\(\gamma\) & weight term in CLIP score\\
\( s_{\square} \) & score to evaluate image quality, footnote $\square$ is the name of the score \\
\(W = \{w_1, w_2, \dots \}\) & weights set for each score measurements \\
\( S \) & weighted score that evaluate multi-aspect image quality \\
\(\Phi\) & VQA Model\\
\bottomrule
    \end{tabular}
\end{table}

\section{Limitations}

Firstly, \modelname{} relies on existing large language models (LLMs) and aesthetic scoring models, whose performance and accuracy could be influenced by the biases and limitations of the LLMs. Secondly, while we introduce random noise to increase image diversity, this method might lead to a reduction in consistency between some images and their text prompts. In addition, due to the high cost of time or money, we have not adopted LLaVA~\cite{liu2023llava}, GPT-4V and latest advanced multimodal large models~\cite{qin2023mp5, liu2023llava, instructblip, zhu2023minigpt, ye2023mplug, zhou2024minedreamer} to generate prompts or QA pairs. In terms of image evaluation, we employ VQA scores, CLIP scores, and aesthetic scores, which may not capture all aspects of image quality.

\section{More Visualized Results}

\begin{figure*}[!ht]
  \centering
    \includegraphics[width=0.8\linewidth]{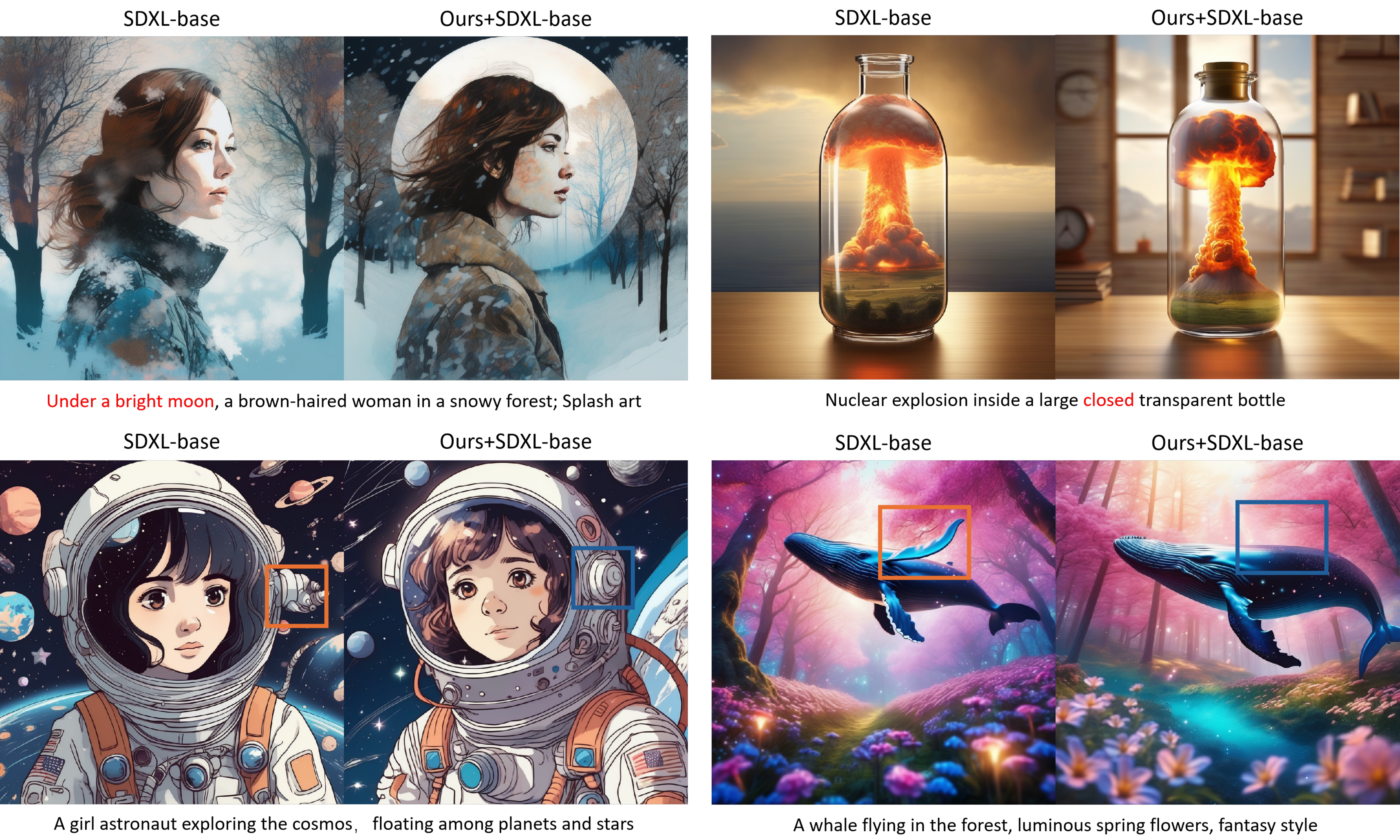}
  \caption{We introduce \modelname{}: a model-agnostic training algorithm that improves text-to-image (T2I) generation models' faithfulness and coherence to text inputs and image aesthetics without human interventions. The images showcase a comparison of the results before and after finetuning SDXL with \modelname{}.}
  \label{fig:motivation}
\end{figure*}

\cref{fig:motivation} showcases the effectiveness of \modelname{} by comparing text-to-image generation results before and after applying the algorithm to SDXL. Through these side-by-side comparisons, we can observe improvements in both prompt faithfulness and image aesthetics, demonstrating how \modelname{} enhances the model's capabilities without requiring human intervention.

\section{Experimental Environments and Settings}

The softwares, more detailed hyperparameters and devices used for sampling and training are displayed in \cref{Setings}.

\begin{table}[!ht]
\centering
\caption{More experimental details of \modelname{}.}
\label{Setings}
\begin{tabular}{C{7cm}|C{10cm}}
\toprule
\textbf{Config} & \textbf{Detail} \\
\midrule
Operating System & Ubuntu 20.04 \\
Python version & 3.10 \\
PyTorch version & 2.2.0 \\
transformers version & 4.37.2 \\
diffusers version & 0.25.0 \\
Number of Inference Steps        & 50        \\
Images per Prompt                & 8         \\
Sampling Precision               & FP16      \\
SDXL-base Resolution                  & $1024\times1024$ \\
SD v1.4 \& SD v1.5 Resolution    & $512\times512$   \\
SDXL-base Batch Size                  & 64        \\
SD v1.4 \& SD v1.5 Batch Size    & 128       \\
Max Training Steps                  & 1000      \\
SDXL-base Learning Rate               & 0.000001  \\
SD v1.4 \& SD v1.5 Learning Rate & 0.0000005 \\
Learning Rate Scheduler          & Cosine    \\
Mixed Precision                  & FP16      \\ 
GPUs for Training                & 8$\times$NVIDIA A100 (80G)     \\
CPUs                             &  Intel(R) Xeon(R) Gold 6248R CPU @ 3.00GHz\\
\bottomrule
\end{tabular}
\end{table}

\section{Efficiency of \modelname{} Illustration}

\begin{table}[!ht]
\centering
\renewcommand{\arraystretch}{1.3} 
\caption{Approximate time consumption of \modelname{} across various stages and models.}
\label{tab:estimated_time}
\begin{tabular}{C{1.7cm}|c|c|c|c|c}
\toprule
\textbf{Model} & \textbf{\begin{tabular}[c]{@{}c@{}}Prompt \&\\ QA Gen.\end{tabular}} & \textbf{Image Gen.} & \textbf{VQA Score}  & \textbf{\begin{tabular}[c]{@{}c@{}}CLIP Score \&\\ Aes. Score\end{tabular}} & \textbf{Training}              \\ \hline
SD v1.4        & \multirow{3}{*}{12h}                                                 & \multirow{2}{*}{6h} & \multirow{2}{*}{7h} & \multirow{3}{*}{44 min}                                                     & \multirow{2}{*}{1.5h} \\ \cline{1-1}
SD v1.5        &                                                                      &                     &                     &                                                                             &                       \\ \cline{1-1} \cline{3-4} \cline{6-6} 
SDXL-base           &                                                                      & 13h                 & 12h                 &                                                                             & 3h                    \\ \bottomrule
\end{tabular}
\end{table}

\cref{tab:estimated_time} presents a comprehensive overview of the time required to apply our methodology \modelname{} across various stages and models, specifically delineating the durations for tasks such as prompt and Q\&A generation, image generation, VQA scoring, combined the CLIP score and aesthetic score, and model training. We compare these processes across SD v1.4, SD v1.5, and SDXL-base models. For prompt and Q\&A generation, all three models require a uniform duration of 12 hours. Image generation and VQA scoring demonstrate variability, with SD v1.4 and SD v1.5 completing in 6 and 7 hours respectively, which contrasts with SDXL-base's longer durations of 13 and 12 hours for these tasks. The evaluation of CLIP and aesthetic scores takes a relatively shorter time, consistently taking 44 minutes across all models. Training times show a distinction between the models, with SD v1.4 and SD v1.5 requiring only 1.5 hours, whereas SDXL-base necessitates a longer commitment of 3 hours. \cref{tab:estimated_time} underscores the efficiency and resource requirements of our method when applied to different models, providing insightful benchmarks for planning and resource allocation.

\section{Discussion on the Marginal Improvement of CLIP Score}

In the ablation study (see in \cref{tab:ablation_experiments}), we observed that the improvement in CLIP score was relatively small. To better understand this, we conducted a thorough analysis and provide a detailed explanation of the marginal improvement in CLIP score.

Firstly, each pair of images in the preference dataset is generated by the same model, which leads to relatively small differences between the two images in each data point. CLIP score is calculated by encoding both the image and the text as embeddings and measuring their similarity, reflecting the overall alignment between the image and the text. This method results in smaller differences in the CLIP score between the two images in the preference dataset. As mentioned in the main text, we scale CLIP score, VQA score, and aesthetic score to a 0-100 range. Statistical analysis of the preference dataset shows that the difference in CLIP score between ``good'' and ``bad'' images is, on average, 5.34, while the differences in VQA score and aesthetic score are 27.49 and 19.48, respectively. Therefore, after training the model, the improvement in CLIP score is somewhat smaller compared to VQA score and aesthetic score.

To further investigate this, we conducted ablation experiments, the results of which are shown in \cref{tab:ablation_experiments}. When using CLIP score alone for evaluation, the improvement is 0.2; when using all three scores (CLIP score, VQA score, and aesthetic score) together, the improvement in CLIP score is 0.3. Notably, when CLIP score is excluded from the evaluation, the improvement is less than 0.1.

Despite the relatively small improvement in CLIP score, it still plays an important role. CLIP score provides a semantic-level evaluation of images and its continuous output compensates for the limitations of VQA score, which is a discrete value. Therefore, even though the improvement in CLIP score is relatively modest, we still consider it an essential part of the evaluation. As shown in our ablation experiments, the inclusion of CLIP score leads to significant improvements in both VQA score and aesthetic score, further validating the importance of CLIP score in multidimensional evaluation.

In conclusion, while the improvement in CLIP score is marginal, it provides additional semantic alignment information in the evaluation process. Especially when combined with VQA and aesthetic scores, it significantly enhances the overall evaluation. Therefore, we believe that CLIP score remains a valuable component of our evaluation methodology.

\section{Discussion on the Weight Selection Method}

In this study, we optimized the weights of CLIP score, VQA score, and aesthetic score through grid search to obtain a weighted score. The selection of these weights is crucial as it directly influences the overall performance of the model and the contribution of each scoring dimension to the final result. We employed the following approach for weight selection:

\subsection{Grid Search Optimization}
To ensure that the total sum of the weights for each score equals 1 and to avoid excessively large or small weights for any score, we set upper and lower bounds for each score's weight and selected specific candidate values within these ranges. This strategy effectively reduces the computational cost of the grid search. Specifically, the candidate values for the weight of CLIP score are $\{0.3, 0.35, 0.4, 0.45, 0.5, 0.55, 0.6\}$, for VQA score are $\{0.3, 0.35, 0.4, 0.45, 0.5, 0.55, 0.6\}$, and for aesthetic score are $\{0.05, 0.1, 0.15, 0.2\}$. We ensured that the sum of the weights for the three scores always equaled 1. Using this approach, we are able to find the optimal combination among multiple possible weight configurations.

\subsection{Reason for the Small Weight of aesthetic Score}
Since the primary goal of this study is to improve the consistency between the generated images and the input text in the image generation model, we found that the weight of the aesthetic score should be relatively small. In the experiments, we also discovered that when the weight of aesthetic score is too large, it may slightly damage the consistency between the image and text. Therefore, during the grid search process, we set the weight range for aesthetic score to be smaller to ensure greater improvement in text-image consistency.

\subsection{Setting Weights for Other Scores}
Both CLIP score and VQA score are closely related to text-image consistency. To emphasize consistency, we set the weights of these two scores higher. Through grid search, we identified a weight configuration for CLIP score, VQA score, and aesthetic score that significantly improved consistency while also enhancing the overall generation quality.

\subsection{Ablation Study Validation}
During the experiments, we conducted an ablation study (see \cref{tab:ablation_experiments}) to validate the effectiveness of different weight configurations. The results show that training with all three scores significantly outperforms training with any single score or a combination of two scores. This further demonstrates the effectiveness of the chosen weight configuration in improving both consistency and generation quality.

In summary, the weight selection is based on the model's goal—improving the consistency between text and image. As such, higher weights are assigned to CLIP score and VQA score, both of which are related to consistency, while aesthetic score was given a relatively lower weight. Through grid search optimization, we ultimately selected the most effective weight combination, which resulted in significant improvements in the generation quality.

\section{Complementary Experiments}

\subsection{SDXL Refiner Preformence in TIFA Benchmark}

We also applied our method to the SDXL+Refiner model, conducting 40 inference steps on the SDXL-base model followed by 10 inference steps using the Refiner model. Utilizing the same random seed, the results on the TIFA benchmark are shown in \cref{tab:sdxl_refiner on tifa}.
\begin{table}[!ht]
\centering
\caption{Results of SDXL+Refiner and \modelname{} (Ours)+SDXL+Refiner for VQA score and aesthetic score on the TIFA benchmark. \textbf{Red} indicates improvement, while \textbf{Green} indicates a decrease. The best scores for each model type are highlighted in \textbf{Bold}. Column ``Sum'' denotes the sum of improvements on \(s_{\text{VQA}}\) and \(s_{\text{Aes.}}\)}
\label{tab:sdxl_refiner on tifa}
\begin{tabular}{llccc}
\toprule
\multicolumn{1}{c}{\textbf{Model}} & \multicolumn{1}{c}{\textbf{Alignment}} & \textbf{\(s_{\text{VQA}}\)} & \textbf{\(s_{\text{Aes.}}\)} & \textbf{Sum} \\ \hline
\multirow{2}{*}{SDXL + Refiner}    & No alignment                           & 82.8         & 61.4         & -            \\
                                   & \modelname{} (Ours)                          & \textbf{83.9}
                                   \textcolor{red}{(+1.1)}   & \textbf{64.1}\textcolor{red}{(+3.7)}   & \textbf{+4.8}         \\ \bottomrule
\end{tabular}
\end{table}

\subsection{Comparison of Win Rates with Draw Thresholds on HPS v2 Benchmark}
\begin{table}[!ht]
\centering
\footnotesize
\caption{Win \& draw rates under gap thresholds of applying \modelname{} on base model vs the original base model on HPS v2.}

\label{tab:threshold_clip_aes}
{
\begin{tabular}{c|c|cc|cc|cc}
\toprule
\multirow{2}{*}{\textbf{\begin{tabular}[c]{@{}c@{}}Tie\\ Threshold\end{tabular}}} & \multirow{2}{*}{\textbf{Score}} & \multicolumn{2}{c|}{\textbf{SD v1.4}} & \multicolumn{2}{c|}{\textbf{SD v1.5}} & \multicolumn{2}{c}{\textbf{SDXL-base}} \\ \cline{3-8} 
                                                                                   &                                 & Win              & Draw              & Win              & Draw              & Win             & Draw           \\ \hline
\multirow{2}{*}{0.1}                                                               & CLIP                            & 52.5\%            & 6.9\%             & 55.4\%            & 4.8\%             & 55.6\%           & 4.3\%          \\ \cline{2-8} 
                                                                                   & Aes.                            & 56.8\%            & 4.9\%             & 63.3\%            & 2.9\%             & 76.3\%           & 1.9\%          \\ \hline
\multirow{2}{*}{0.01}                                                              & CLIP                            & 56.0\%            & 0.6\%             & 57.8\%            & 0.4\%             & 57.6\%           & 0.3\%          \\ \cline{2-8} 
                                                                                   & Aes.                            & 59.1\%            & 0.6\%             & 64.3\%            & 0.3\%             & 77.1\%           & 0.3\%          \\ \hline
\multirow{2}{*}{0.001}                                                             & CLIP                            & 56.2\%            & 0.1\%             & 57.9\%            & 0.03\%            & 57.7\%           & 0.0\%          \\ \cline{2-8} 
                                                                                   & Aes.                            & 59.2\%            & 0.03\%            & 64.6\%            & 0.0\%            & 77.3\%           & 0.05\%          \\
\bottomrule
\end{tabular}
}

\end{table}

In the HPS v2 benchmark test, we establish a gap threshold (0.1, 0.01, 0.001) to determine the outcomes of comparisons. Results are considered a draw when the absolute value of the gap is less than or equal to this threshold. In \cref{tab:threshold_clip_aes}, we present the win rates and probabilities of drawing at different gap thresholds for the SD v1.4, SD v1.5, and SDXL-base models trained using \modelname{}, compared with the original models, across both CLIP scores and aesthetic scores. The results clearly demonstrate that the models finetuned with \modelname{} consistently outperform the base models. Moreover, even when adjusting the gap threshold to 0.1, the changes in win rates remain minimal.

\subsection{Analysis of Prompts Filtering for High-Quality Image Generation}

We also employ the method described in ``DreamSync" to filter prompts capable of generating high-quality images from our generated prompts. Specifically, we filter prompts that can generate images with a VQA score \textgreater 0.9 and an aesthetic score \textgreater 0.6 using the SD v1.5 model. \cref{tab:prompts_questions_stats_updated} presents the attributes of the text and the questions generated from the filtered prompts, while \cref{tab:category_numbers_proportions} displays the number and proportion of each category of prompts obtained through filtering. The results indicate that, although there is no significant change in the nature of the text and the generated questions of the filtered prompts, the proportion of filtered prompts varies greatly across different categories, with the difference in filtering proportion reaching up to 33.5\%. This suggests that the types of prompts that the SD v1.5 model, or T2I models in general, excel at vary significantly across categories. Merely selecting prompts capable of producing high-quality images for training is insufficient for a comprehensive approach.

\begin{table}[!ht]
\centering

\caption{In our dataset, statistics of prompts that can generate images with VQA score \textgreater 0.9 and aesthetic score \textgreater 0.6 through the SD v1.5 model.}
\label{tab:prompts_questions_stats_updated}
\begin{tabular}{l|c}
\toprule
\multicolumn{1}{c|}{\textbf{Statistic}} & \textbf{Value} \\
\midrule
Total number of prompts                    & 15,262         \\
Total number of questions                      & 132,893        \\
\midrule
Average number of questions per prompt         & 8.71           \\
Average number of words per prompt            & 26.75          \\
Average number of elements in prompts          & 8.05           \\
Average number of words per question           & 7.94           \\
\bottomrule
\end{tabular}
\end{table}

\begin{table}[!ht]
\centering
\caption{In our dataset, the number of various categories of prompts that can generate images with VQA score \textgreater 0.9 and aesthetic score \textgreater 0.6 through the SD v1.5 model, and their retention proportions compared to the original categories of prompts.}
\label{tab:category_numbers_proportions}
\begin{tabular}{ccc}
\toprule
\textbf{Category} & \textbf{Count} & \textbf{     Retention Proportion      } \\
\midrule
Natural Landscapes & 1992 & 34.7\% \\
Cities and Architecture & 2046 & 32.5\% \\
People & 1950 & 38.7\% \\
Animals & 1347 & 43.6\% \\
Plants & 1849 & 43.2\% \\
Food and Beverages & 1116 & 37.1\% \\
Sports and Fitness & 1060 & 35.4\% \\
Art and Culture & 714 & 29.4\% \\
Technology and Industry & 853 & 26.5\% \\
Everyday Objects & 712 & 26.1\% \\
Transportation & 1362 & 30.6\% \\
Abstract and Conceptual & 261 & 10.1\% \\
\bottomrule
\end{tabular}
\end{table}

\subsection{Results of Training Models with Different Amounts of Data}

Our preference dataset construction method achieves a 100\% data conversion efficiency, highlighting the importance of high data conversion efficiency. To further demonstrate this, we randomly select different proportions of data from our dataset for experiments. Moreover, following the strategy introduced in "DreamSync", we filter prompts capable of generating high-quality images based on two sets of thresholds (VQA score \textgreater 0.85 and aesthetic score \textgreater 0.5, and VQA score \textgreater 0.9 and aesthetic score \textgreater 0.6) to construct a preference dataset for training, using SD v1.5 as the base model and conducting the experiment on the TIFA benchmark. The experimental results are shown in \cref{tab:different_rate_data_train_result}. The results indicate that although the aesthetic scores surpass the scenario of using all data when trained with 60\% and 80\% of the data, considering the CLIP score, VQA score, and aesthetic score together, the larger the volume of data, the better the overall performance of the model. Especially, the improvement in model performance when increasing data usage from 60\% to 100\% is significantly greater than that from 0\% to 60\%. Notably, with only our limited constructed preference dataset (i.e. 20\% of the preference dataset), we have significant improvements in all aspects, further demonstrating \modelname{}'s efficiency.

\begin{table}[!ht]
\caption{Evaluation results of training the SD v1.5 model with different amounts of data on the TIFA benchmark. The second column shows the method used for data selection. Red indicates improvement relative to the original SD v1.5, while green indicates a decrease. We highlight the highest score in each column in bold.}
\centering
\renewcommand{\arraystretch}{1.2}
\begin{tabular}{ccC{2cm}C{2cm}C{2cm}}
\toprule

\textbf{Proportion} & \textbf{Sample Method}                                   & \multicolumn{1}{c}{\textbf{\(s_{\text{CLIP}}\)}} & \multicolumn{1}{c}{\textbf{\(s_{\text{VQA}}\)}} & \multicolumn{1}{c}{\textbf{\(s_{\text{Aes.}}\)}} \\ \hline
0                                   & -                                                        & 27.0                                    & 77.1                                   & 48.0                                         \\ \hline
20\%                                & Random Sample                                            & 27.2\textcolor{red}{(+0.2)}                              & 77.4\textcolor{red}{(+0.3)}                             & 49.1\textcolor{red}{(+1.1)}                                   \\
33.3\%                              & $s_{VQA}>0.9,s_{Aes.}>0.6$ & 27.1\textcolor{red}{(+0.1)}                              & 77.4\textcolor{red}{(+0.3)}                             & 49.1\textcolor{red}{(+1.1)}                                   \\
40\%                                & Random Sample                                            & 27.1\textcolor{red}{(+0.1)}                              & 77.5\textcolor{red}{(+0.4)}                             & 48.8\textcolor{red}{(+0.8)}                                   \\
60\%                                & Random Sample                                            & 27.1\textcolor{red}{(+0.1)}                              & 77.6\textcolor{red}{(+0.5)}                             & \textBF{49.3}\textcolor{red}{(+1.3)}                                   \\
60.6\%                              & $s_{VQA}>0.85,s_{Aes.}>0.5$                            & \textBF{27.3}\textcolor{red}{(+0.3)}                              & 77.7\textcolor{red}{(+0.6)}                             & 47.6\textcolor{green!60!black}{($-$0.4)}                                   \\
80\%                                & Random Sample                                            & \textBF{27.3}\textcolor{red}{(+0.3)}                              & 78.3(\textcolor{red}{+1.2)}                             & 49.2\textcolor{red}{(+1.2)}                                   \\ \hline
100\%                               & -                                                        & \textBF{27.3}\textcolor{red}{(+0.3)}                              & \textBF{78.7}\textcolor{red}{(+1.6)}                             & 49.1\textcolor{red}{(+1.1)}                                   \\ \bottomrule

\label{tab:different_rate_data_train_result}
\end{tabular}

\end{table}

\subsection{Prompt Utilization Rate Analysis}

\begin{figure*}[htbp]
  \centering
  \begin{subfigure}[b]{0.48\textwidth}
  \centering
  
    \includegraphics[width=0.7\textwidth]{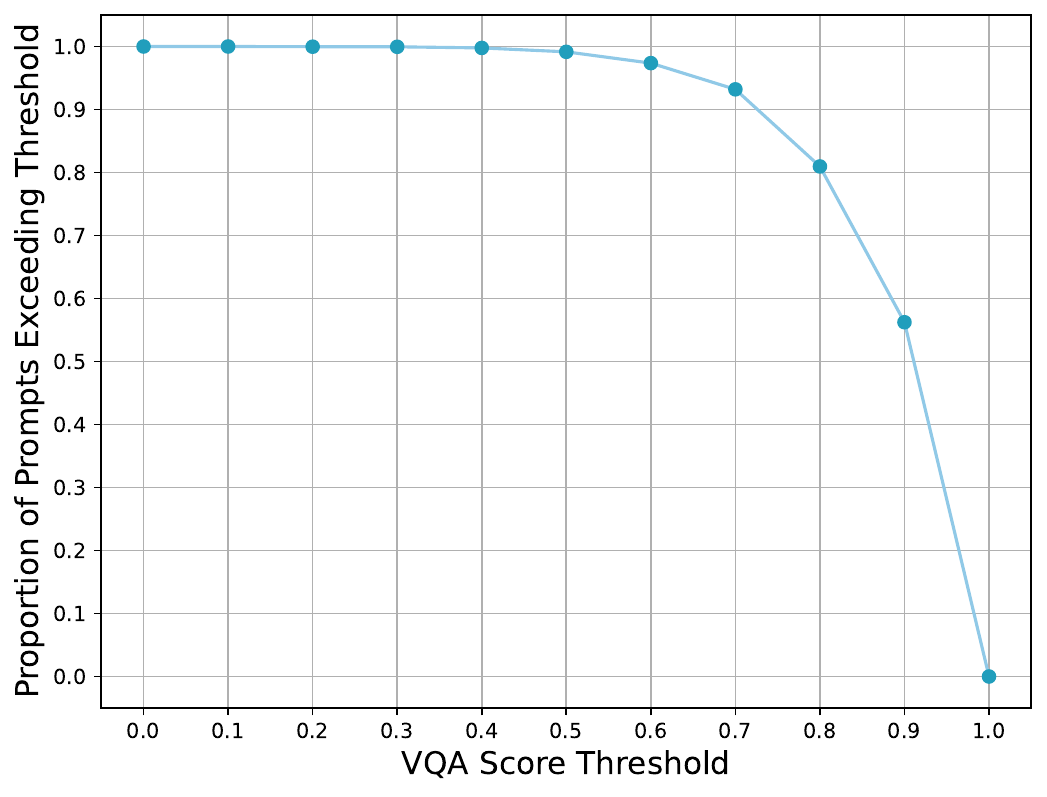}
    \label{fig:vqa_score_fliter}
  \end{subfigure}
  \hfill 
  \begin{subfigure}[b]{0.48\textwidth}
  \centering
  
    \includegraphics[width=0.7\textwidth]{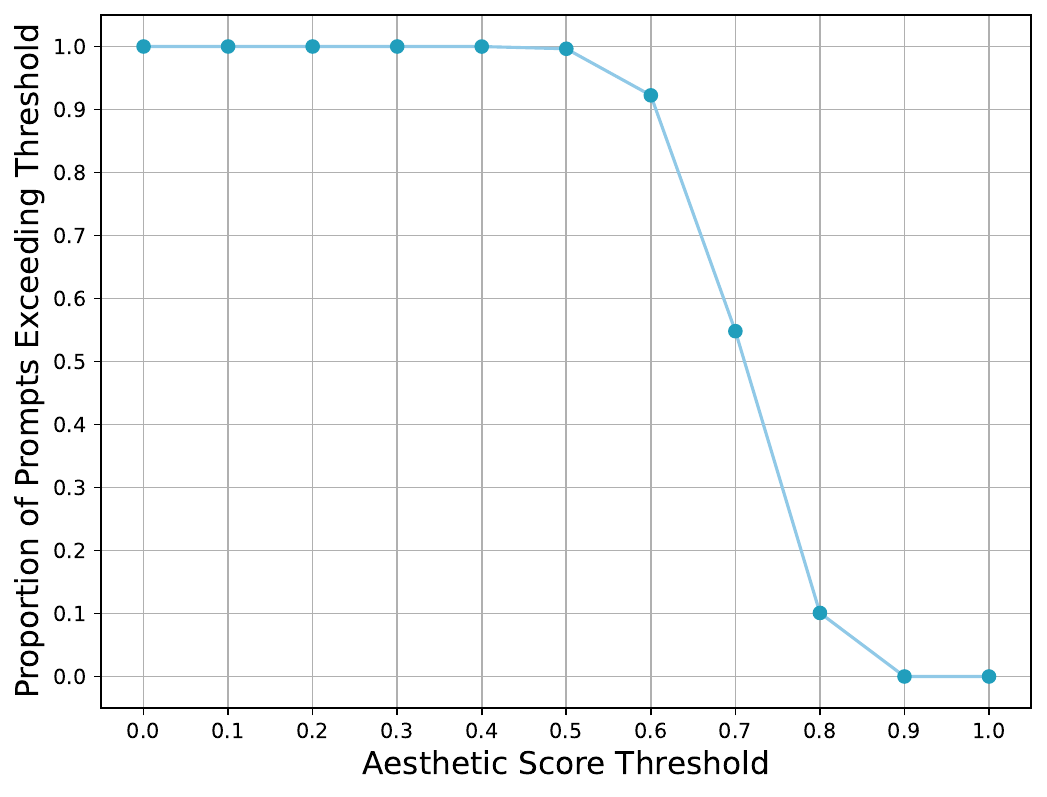}
    \label{fig:aes_score_fliter}
  \end{subfigure}
  \caption{In our generated dataset constructed with images generated by SDXL-base, the proportion of prompts that can be filtered out based on varying thresholds.}
  \label{fig:vqa_aes_score_fliter}
\end{figure*}

To compare with DreamSync~\cite{sun2023dreamsync}, in \cref{fig:vqa_aes_score_fliter}, we present the number of prompts that can be selected from our generated dataset as the thresholds for VQA score and aesthetic score vary. Specifically, these are prompts for which at least one generated image scores above the respective thresholds. With DreamSync's 0.9 and 0.6 thresholds for two metrics, only 48.8\% of prompts satisfying both, meaning a low data conversion efficiency of 48.8\% on our dataset. However, \modelname{}'s approach merely requires selecting the best and worst images without imposing any threshold constraints, thereby achieving a data conversion efficiency of 100\%.

\section{LLM and VLM Instructions Details}

\subsection{Example Instruction to Generate Image Caption}

Using LLM to generate image captions is the first step in constructing a dataset. In this step, we use GPT-3.5 to generate diverse image captions. When generating captions, we specify the category of the caption and provide five examples. Below is an example of an instruction:

\textit{
You are a large language model, trained on a massive dataset of text. You can generate texts from given examples. You are asked to generate similar examples to the provided ones and follow these rules: }
\begin{enumerate}

\item \textit{Your generation will be served as prompts for Text-to-Image models. So your prompt should be as visual as possible.}

\item \textit{Do NOT generate scary prompts. }

\item \textit{Do NOT repeat any existing examples. }

\item \textit{Your generated examples should be as creative as possible. }

\item \textit{Your generated examples should not have repetition. }

\item \textit{Your generated examples should be as diverse as possible.} 

\item \textit{Do NOT include extra texts such as greetings.}

\item \textit{Generate \{num\} descriptions.}

\item \textit{The descriptions you generate should have a diverse word count, with both long and short lengths.}

\item \textit{The more detailed the description of an image, the better, and the more elements, the better.}
\end{enumerate}

\textit{Please open your mind based on the theme "Natural Landscapes: Includes terrain, bodies of water, weather phenomena, and natural scenes." paintings}

\textit{Here are five example descriptions for natural landscape images:}
\begin{enumerate}

\item \textit{A sprawling meadow under a twilight sky, where the last rays of the sun kiss the tips of wildflowers, creating a canvas of gold and purple hues.}

\item \textit{A majestic waterfall cascading down rugged cliffs, enveloped by a mist that dances in the air, surrounded by an ancient forest whispering the tales of nature.}

\item \textit{An endless desert, where golden dunes rise and fall like waves in an ocean of sand, punctuated by the occasional resilient cactus standing as a testament to life's perseverance.}

\item \textit{A serene lake, mirror-like, reflecting the perfect image of surrounding snow-capped mountains, while a solitary swan glides gracefully, leaving ripples in its wake.}

\item \textit{The aurora borealis illuminating the polar sky in a symphony of greens and purples, arching over a silent, frozen landscape that sleeps under a blanket of snow.}
\end{enumerate}

\textit{Please imitate the example above to generate a diverse image description and do not repeat the example above.}

\textit{Each description aims to vividly convey the beauty and unique atmosphere of various natural landscapes.}

\textit{The format of your answer should be: }
\begin{lstlisting}
{
    "descriptions":[...]
}
\end{lstlisting}
\textit{Ensure that the response can be parsed by} \verb|json.loads| \textit{in Python, for example: no trailing commas, no single quotes, and so on.}

\subsection{Instruction to Generate Question and Answer Pairs with Validation}

After obtaining a large number of image captions, we also need to break these captions down into Question-Answer (QA) pairs. For this step, we use Gemini Pro, requesting it to decompose each image caption into 15 QA pairs, with each caption processed six times. Finally, we filter out the QA pairs that are generated repeatedly. Below is the instruction given to Gemini Pro for breaking down image captions into QA pairs.

\textit{You are a large language model, trained on a massive dataset of text. You can receive the text as a prompt for Text-to-Image models and break it down into general interrogative sentences that verifies if the image description is correct and give answers to those questions.}

\textit{You must follow these rules:}
\begin{enumerate}
\item \textit{Based on the text content, the answers to the questions you generate must only be 'yes', meaning the questions you generate should be general interrogative sentences.}

\item \textit{The questions you generate must have a definitive and correct answer that can be found in the given text, and this answer must be 'yes'.}

\item \textit{The correct answer to your generated question cannot be unmentioned in the text, nor can it be inferred solely from common sense; it must be explicitly stated in the text.}

\item \textit{Each question you break down from the text must be unique, meaning that each question must be different.}

\item \textit{If you break down the text into questions, each question must be atomic, i.e., they must not be divided into new sub-questions.}

\item \textit{Categorize each question into types (object, human, animal, food, activity, attribute, counting, color, material, spatial, location, shape, other).}

\item \textit{You must generate at least 15 questions, ensuring there are at least 15 question ids.}

\item \textit{The questions you generate must cover the content contained in the text as much as possible.}

\item \textit{You also need to indicate whether the question you provided is an invalid question of the "not mentioned in the text" type, with 0 representing an invalid question and 1 representing a minor question.}
\end{enumerate}

\textit{Each time I'll give you a text that will serve as a prompt for Text-to-Image models.}

\textit{You should only respond in JSON format as described below:}
\begin{lstlisting}
[
    {
        "question_id": "The number of the issue you generated, starting with 1",
        "question": "A general interrogative sentence you derive from breaking down the text should inquire whether the image conforms to the content of the text. The answer to this question must be found based on the text, not on common sense, etc. The answer must not be unmentioned in the text, and according to the text, the answer to this question must be 'yes'.",
        "answer": "The real answer to the question according to the text provided. The answer should be 'yes'",
        "element_type": "The type of problem. (object, human, animal, food, activity, attribute, counting, color, material, spatial, location, shape, other)",
        "element": "The elements mentioned in the question, or the specific elements asked by the question",
        "flag": "Check if the correct answer to the question you generated is an invalid question such as not mentioned, with 0 being an invalid question and 1 being not an invalid question"
    }
    # There should be more questions here, because a text should be broken down into multiple questions, and the number of questions is up to you
]
\end{lstlisting}
\textit{Ensure that the response can be parsed by} \verb|json.loads| \textit{in Python, for example: no trailing commas, no single quotes, and so on.}

\subsection{Detailed Instruction When Using GPT-4V for Evaluation}

When evaluating the model trained with our method using GPT-4V, to allow GPT-4V to decide whether the image generated by the post-training model is better or the one generated by the original model is better, we designed the following instruction:

\textit{The prompt for these two pictures is: \{prompt\}
Which image do you prefer? No matter what happens, you must make a choice and answer A or B. }

\textit{Reply in JSON format below:}
\begin{lstlisting}
{
  "reason": "your reason",
  "choice": "A/B"
}
\end{lstlisting}

\textit{Which image better fits the text description? No matter what happens, you must make a choice and answer A or B. }

\textit{Reply in JSON format below:}
\begin{lstlisting}
{
  "reason": "your reason",
  "choice": "A/B"
}
\end{lstlisting}

\textit{Disregarding the prompt, which image is more visually appealing? No matter what happens, you must make a choice and answer A or B. }

\textit{Reply in JSON format below:}
\begin{lstlisting}
{
  "reason": "your reason",
  "choice": "A/B"
}
\end{lstlisting}

\section{Details of Generated Prompts and Preference Dataset}

\subsection{Statistics of the Prompts for Different Categories}

The statistics of our LLM-generated prompts for preference candidate sets' image captions, comprising 45,834 prompts, are presented in Table~\ref{tab:generated_dataset_stats} across 12 distinct categories: Natural Landscapes, Cities and Architecture, People, Animals, Plants, Food and Beverages, Sports and Fitness, Art and Culture, Technology and Industry, Everyday Objects, Transportation, and Abstract and Conceptual Art.

\begin{table}[htbp]
    \centering
    \caption{Distribution of prompts across different categories in our dataset.}
    \label{tab:generated_dataset_stats}
    \begin{tabular}{lc}
        \toprule
        \textbf{Category} & \textbf{Count} \\
        \midrule
        Natural Landscapes & 5,733 \\
        Cities and Architecture & 6,291 \\
        People & 5,035 \\
        Animals & 3,089 \\
        Plants & 4,276 \\
        Food and Beverages & 3,010 \\
        Sports and Fitness & 2,994 \\
        Art and Culture & 2,432 \\
        Technology and Industry & 3,224 \\
        Everyday Objects & 2,725 \\
        Transportation & 4,450 \\
        Abstract and Conceptual Art & 2,575 \\
        \midrule
        Total & 45,834 \\
        \bottomrule
    \end{tabular}
\end{table}

As shown in Table~\ref{tab:generated_dataset_stats}, the distribution of prompts across categories varies, with Cities and Architecture having the highest count (6,291) and Art and Culture having the lowest (2,432). This diversity in prompt distribution ensures a wide range of concepts and subjects for our T2I model to learn from and generate images.

\subsection{Statistics of the AI-Generated Captions}

\cref{tab:dataset_statistic} presents more detailed data of our generated dataset. In addition to this, in our generated QA pairs, the counts for each category are as follows: shape (2385), counting (3809), material (4495), food (4660), animal (5533), color (12749), human (17921), spatial (21878), other (24513), location (42914), object (77783), activity (83712), and attribute (101713).

\begin{table}[!ht]
\centering
\caption{Summary statistics of QA pair dataset}
\label{tab:dataset_statistic}
\begin{tabular}{l|c}
\toprule
 
\multicolumn{1}{c|}{\textbf{Statistic}}                          & \textbf{Value}   \\

\midrule
Total number of prompts                       & 45,834           \\

Total number of questions                     & 414,172          \\ \midrule

Average number of questions per prompt        & 9.03             \\

Average number of words per prompt            & 26.061           \\

Average number of elements in prompts         & 8.22             \\

Average number of words per question          & 8.07             \\
\bottomrule
\end{tabular}
\end{table}

\subsection{Example Generated Image Caption and Corresponding QA Pairs}

For the example prompt: \textit{A vast, open savannah, where golden grasses sway in the wind, dotted with acacia trees and herds of majestic elephants and giraffes, as the sun sets  on the horizon.} We have the corresponding QA pairs as in \cref{tab:corresponding_qa_pairs}. As indicated in \cref{tab:corresponding_qa_pairs}, for each generated question, we require the LLM to provide the main element and type of element being asked in the question.

\begin{table}[!ht]

\caption{Example of corresponding QA pairs.}
\label{tab:corresponding_qa_pairs}
\centering
\begin{tabular}{c|c|c}
\toprule
\textbf{Question and Choices} & \textbf{Type} & \textbf{Element} \\
\midrule
\makecell[l]{Q: Is there a vast, open savanna in the image? \\ A: Yes} & location & savannah \\ \midrule
\makecell[l]{Q: Is there golden grass in the savannah? \\ A: Yes} & object & golden grass \\ \midrule
\makecell[l]{Q: Do golden grasses sway in the wind \\in the described scene? \\ A: Yes} & activity & golden grasses \\ \midrule
\makecell[l]{Q: Is there a mention of acacia trees in the image? \\ A: Yes} & object & acacia trees \\ \midrule
\makecell[l]{Q: Are there majestic elephants in the savannah? \\ A: Yes} & animal & elephants \\ \midrule
\makecell[l]{Q: Are the giraffes majestic? \\ A: Yes} & attribute & giraffes \\ \midrule
\makecell[l]{Q: Is there a sun setting on the horizon? \\ A: Yes} & activity & sun setting \\
\bottomrule
\end{tabular}
\end{table}

\subsection{Example Preference Dataset Generated by \modelname{}}

\begin{figure}[!ht]
\centering
\includegraphics[width=0.9\textwidth]{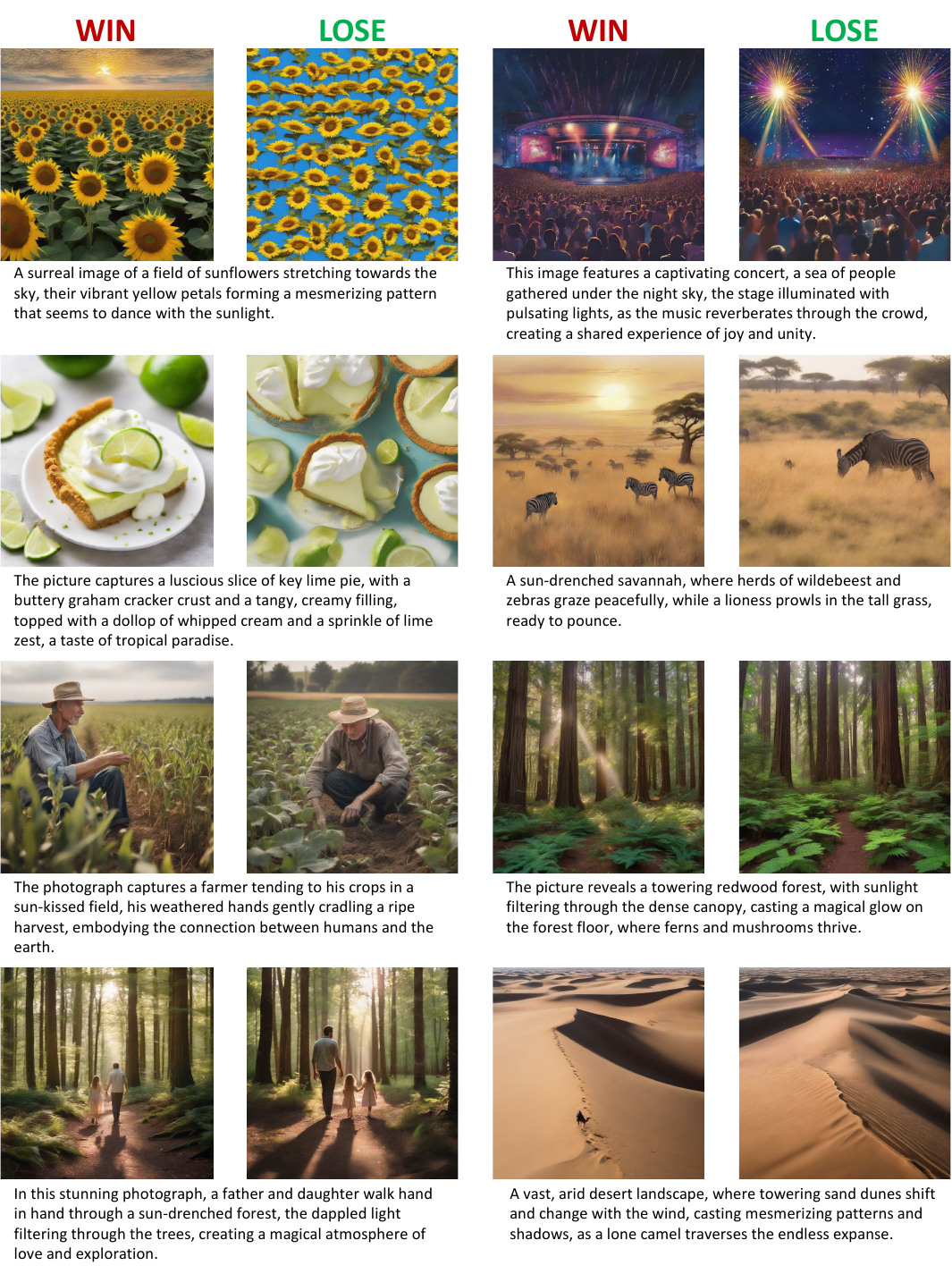}
\caption{Some examples of the preference dataset we generated.}
\label{example dataset}
\end{figure}

As shown in \cref{example dataset}, each data set consists of a high-quality image, a lower-quality image, and a corresponding image caption. 

\end{document}